\documentclass[english]{article}

\usepackage{times}
\usepackage{a4wide}
\usepackage{graphicx} 
\usepackage{subfigure} 

\usepackage{natbib}

\usepackage{algorithm}
\usepackage{algorithmic}

\usepackage{hyperref}
\usepackage{url}
\usepackage{graphicx}
\usepackage{amsmath}
\usepackage{amssymb}
\usepackage{bm}
\usepackage{dsfont}
\usepackage{wrapfig}

\usepackage{todonotes}

\author{
Daniel Hern\'andez-Lobato\\
Universidad Aut\'onoma de Madrid\\
Francisco Tom\'as y Valiente 11\\
28049, Madrid, Spain\\
\texttt{daniel.hernandez@uam.com} \\
\and
Jos\'e Miguel Hern\'andez-Lobato\\
Harvard University\\
33 Oxford street\\
Cambridge, MA 02138, USA\\
\texttt{jmhl@seas.harvard.edu} \\
\and
Amar Shah\\
Cambridge University\\
Trumpington Street, Cambridge\\
CB2 1PZ, United Kingdom.\\
\texttt{as793@cam.ac.uk} \\
\and 
Ryan P. Adams\\
Harvard University and Twitter\\
33 Oxford street\\
Cambridge, MA 02138, USA\\
\texttt{rpa@seas.harvard.edu} \\
}

\date{}

\title{Predictive Entropy Search for Multi-objective Bayesian Optimization}

\begin{document} 

\maketitle

\begin{abstract} 
We present {\small PESMO}, a Bayesian method for identifying the Pareto set of multi-objective 
optimization problems, when the functions are expensive to evaluate.
The central idea of {\small PESMO} is to choose evaluation points so as to maximally reduce 
the entropy of the posterior distribution over the Pareto set. Critically, the {\small PESMO} 
multi-objective acquisition function can be decomposed as a sum of objective-specific acquisition 
functions, which enables the algorithm to be used in \emph{decoupled} scenarios in 
which the objectives can be evaluated separately and perhaps with different costs. This decoupling 
capability also makes it possible to identify difficult objectives that require more evaluations.  
{\small PESMO} also offers gains in efficiency, as its cost scales linearly with the number of objectives, 
in comparison to the exponential cost of other methods. We compare {\small PESMO} with other related methods for multi-objective 
Bayesian optimization on synthetic and real-world problems. The results show that {\small PESMO} produces better recommendations with 
a smaller number of evaluations of the objectives, and that a decoupled evaluation can lead to improvements 
in performance, particularly when the number of objectives is large.
\end{abstract} 

\section{Introduction}

We address the problem of optimizing $K$ real-valued functions~${f_1(\mathbf{x}),\ldots,f_K(\mathbf{x})}$ over some bounded domain~${\mathcal{X}\subset \mathds{R}^d}$, where $d$ is the dimensionality of the input space. 
This is a more general, challenging and realistic scenario than the one considered in traditional optimization problems where there is a single-objective function.
For example, in a complex robotic system, we may be interested in minimizing the energy consumption while maximizing locomotion speed \cite{AriizumiTCM14}.  When selecting a financial portfolio, it may be desirable to maximize returns while minimizing various risks.  In a mechanical design, one may wish to minimize manufacturing cost while maximizing durability.
In each of these multi-objective examples, it is unlikely to be possible to optimize all of the objectives simultaneously as they may be conflicting: a fast-moving robot probably consumes more energy, high-return financial instruments typically carry greater risk, and cheaply manufactured goods are often more likely to break.
Nevertheless, it is still possible to find a set of optimal points $\mathcal{X}^\star$ known as the \emph{Pareto set} \cite{collette2003}.
Rather than a single best point, this set represents a collection of solutions at which no objective can be improved without damaging one of the others.

In the context of minimization, we say that $\mathbf{x}$ \emph{Pareto dominates} $\mathbf{x}'$ if~${f_k(\mathbf{x}) \leq f_k(\mathbf{x}')\;\forall k}$, with at least one of the inequalities being strict.
The \emph{Pareto set}~$\mathcal{X}^\star$ is then the subset of non-dominated points in~$\mathcal{X}$, i.e., the set such that ${\forall \mathbf{x}^\star \in \mathcal{X}^\star}$, ${\forall \mathbf{x} \in \mathcal{X}}$, ${\exists\, k \in 1,\ldots,K}$ for which~${f_k(\mathbf{x}^\star) < f_k(\mathbf{x})}$.
The Pareto set is considered to be optimal because for each point in that set one cannot improve in one of the objectives without deteriorating some other objective.
Given~$\mathcal{X}^\star$, the user may choose a point from this set according to their preferences, \emph{e.g.}, locomotion speed vs.\ energy consumption.
The Pareto set is often not finite, and most strategies aim at finding a finite set with which to approximate $\mathcal{X}^\star$ well.

It frequently happens that there is a high cost to evaluating one or more of the functions $f_k(\cdot)$.
For example, in the robotic example, the evaluation process may  involve a time consuming experiment with the embodied robot.
In this case, one wishes to minimize the number of evaluations required to obtain a useful  approximation to the Pareto set~$\mathcal{X}^\star$.
Furthermore, it is often the case that there is no simple closed form for the objectives~$f_k(\cdot)$, \emph{i.e.}, they can be regarded as black boxes.
One promising approach in this setting has been to use a probabilistic model such as a Gaussian process to approximate each function \cite{knowles2006,Emm08,Ponweiser2008,Picheny2015}.
At each iteration, these strategies  use the uncertainty captured by the probabilistic model to generate an acquisition (utility) function, the maximum of which provides an effective heuristic for identifying a promising location on which to evaluate the objectives.
Unlike the actual objectives, the acquisition function is a function of the model and therefore relatively cheap to evaluate and maximize.
This approach contrasts with model-free methods based on genetic algorithms or  evolutionary strategies that are known to be effective for approximating the Pareto set, but demand a large number of function evaluations \cite{deb2002,Li2003,Zitzler99}. 

Despite these successes, there are notable limitations to current model-based approaches: 1) they often build the acquisition function by transforming the multi-objective problem into a single-objective problem using scalarization  techniques (an approach that is expected to be suboptimal), 2) the acquisition function generally requires the evaluation of \emph{all} of the objective functions at the same location in each iteration, and 3) the computational cost of evaluating the acquisition function typically grows exponentially with the number of objectives, which limits their applicability to optimization problems with just 2 or 3 objectives.

We describe here a strategy for multi-objective optimization that addresses these concerns.
We extend previous single-objective strategies based on stepwise uncertainty reduction to the multi-objective 
case \cite{Villemonteix2009,NIPS2014_5324,HennigSchuler2012}.
In the single-objective case, these strategies choose the next evaluation location based on the reduction 
of the Shannon entropy of the posterior estimate of the minimizer~$\mathbf{x}^\star$.
The idea is that a smaller entropy implies that the minimizer $\mathbf{x}^\star$ is better identified; the heuristic then 
chooses candidate evaluations based on how much they are expected to improve the quality of this estimate.
These information gain criteria have been shown to often provide better results than other 
alternatives based, \emph{e.g.}, on the popular \emph{expected improvement}  \cite{NIPS2014_5324,HennigSchuler2012,NIPS2015_5804}. 

The extension to the multi-objective case is obtained by considering the entropy of the 
posterior distribution over the Pareto set~$\mathcal{X}^\star$.
More precisely, we choose the next evaluation as the one that is expected to most reduce the entropy of our estimate of $\mathcal{X}^\star$.
The proposed approach is called \emph{predictive entropy search for multi-objective optimization} ({\small PESMO}).
Several experiments involving real-world and synthetic optimization problems, show that {\small PESMO} can lead to better 
performance than related methods from the literature.
Furthermore, in {\small PESMO} the acquisition function is expressed as a sum across the different objectives, allowing for \emph{decoupled} scenarios in which we can choose to only evaluate a subset of objectives at any given location.
In the robotics example, one might be able to decouple the problems by estimating energy consumption from a simulator even if the locomotion speed could only be evaluated via physical experimentation.
Another example, inspired by \citet{gelbart2014}, might be the design of a low-calorie cookie: one wishes to maximize taste while minimizing calories, but calories are a simple function of the ingredients, while taste could require human trials.
The results obtained show that {\small PESMO} can 
obtain better results with a smaller number of evaluations of the objective functions in such scenarios.
Furthermore, we have observed that the decoupled evaluation provides significant improvements over a coupled evaluation when the number of objectives is large. Finally, unlike other methods \cite{Ponweiser2008,Picheny2015}, the computational 
cost of {\small PESMO} grows linearly with the number of objectives.


\section{Multi-objective Bayesian Optimization via Predictive Entropy Search}
\label{sec:pesm}

In this section we describe the proposed approach for multi-objective optimization  based on \emph{predictive entropy search}.
Given some previous evaluations of each objective function~$f_k(\cdot)$, we seek to choose new evaluations that maximize the information gained about the Pareto set~$\mathcal{X}^\star$.
This approach requires a probabilistic model for the unknown objectives, and we therefore assume that each~$f_k(\cdot)$ follows a Gaussian process (GP) prior \cite{rasmussen2005book}, with observation noise that is i.i.d.\ Gaussian with zero mean.
GPs are often used in model-based approaches to multi-objective optimization because of their flexibility and ability to model uncertainty \cite{knowles2006,Emm08,Ponweiser2008,Picheny2015}.
For simplicity, we initially consider a coupled setting in which we evaluate all objectives at the same location in any given iteration.
Nevertheless, the approach described can be easily extended to the \emph{decoupled} scenario.

Let ${\mathcal{D} = \{(\mathbf{x}_n,\mathbf{y}_n)\}_{n=1}^N}$ be the data (function evaluations) collected up to step~$N$, where~$\mathbf{y}_n$ is a $K$-dimensional vector with the values resulting from the evaluation of all objectives at step $n$, and~$\mathbf{x}_n$ is a vector in input space denoting the evaluation location.
The next query~$\mathbf{x}_{N+1}$ is the one that  maximizes the expected reduction in the entropy~$H(\cdot)$ of the posterior distribution over the Pareto set~$\mathcal{X}^\star$, i.e., $p(\mathcal{X}^\star|\mathcal{D})$.
The acquisition function of {\small PESMO} is hence:
\begin{align}
\alpha(\mathbf{x}) &= H(\mathcal{X}^\star|\mathcal{D}) - 
\mathds{E}_\mathbf{y}\left[ H(\mathcal{X}^\star|\mathcal{D} \cup \{(\mathbf{x}, \mathbf{y})\})\right]\,, 
\label{eq:exact_acq}
\end{align} 
where $\mathbf{y}$ is the output of all the GP models at~$\mathbf{x}$ and the expectation is taken with respect to the posterior distribution for~$\mathbf{y}$ given by these models,~${p(\mathbf{y}|\mathcal{D},\mathbf{x}) = \prod_{k=1}^K p(y_k|\mathcal{D},\mathbf{x})}$.
The GPs are assumed to be independent \emph{a priori}.
This acquisition function is known as \emph{entropy search} \citep{Villemonteix2009,HennigSchuler2012}.
Thus, at each iteration we set the location of the next evaluation to~${\mathbf{x}_{N+1}=\text{arg max}_{\mathbf{x} \in \mathcal{X}} \, \alpha(\mathbf{x})}$.

A practical difficulty, however, is that the exact evaluation of Eq.~(\ref{eq:exact_acq}) is generally infeasible and the function must be approximated; we follow the approach described in \citet{NIPS2014_5324,Houlsby2012}.
In particular, Eq.~(\ref{eq:exact_acq}) is the mutual information between~$\mathcal{X}^\star$ and~$\mathbf{y}$ given~$\mathcal{D}$.
The mutual information is symmetric and hence we can exchange the roles of the variables~$\mathcal{X}^\star$ and~$\mathbf{y}$, leading to an expression that is equivalent to Eq.~(\ref{eq:exact_acq}):
\begin{align}
\alpha(\mathbf{x}) &= H(\mathbf{y}|\mathcal{D},\mathbf{x}) - 
\mathds{E}_{\mathcal{X}^\star}\left[ H(\mathbf{y}|\mathcal{D},\mathbf{x}, \mathcal{X}^\star)\right]\,,
\label{eq:exact_acq_rev}
\end{align}
where the expectation is now with respect to the posterior distribution for the Pareto set $\mathcal{X}^\star$ given the observed data, and~$H(\mathbf{y}|\mathcal{D},\mathbf{x}, \mathcal{X}^\star)$ measures the entropy of~$p(\mathbf{y}|\mathcal{D},\mathbf{x}, \mathcal{X}^\star)$, \emph{i.e.}, the predictive distribution for the objectives at $\mathbf{x}$ given $\mathcal{D}$ and conditioned to $\mathcal{X}^\star$ being the Pareto set of the objective functions.
This alternative formulation is known as \emph{predictive entropy search} \cite{NIPS2014_5324} and it significantly simplifies the evaluation of the acquisition function~$\alpha(\cdot)$.
In particular, we no longer have to evaluate or approximate the entropy of the Pareto set,~$\mathcal{X}^\star$, which may be quite difficult.
The new acquisition function obtained in Eq.~(\ref{eq:exact_acq_rev}) favors the evaluation in the 
regions of the input space for which~$\mathcal{X}^\star$ is more informative about $\mathbf{y}$.
These  are precisely also the regions in which~$\mathbf{y}$ is more informative about~$\mathcal{X}^\star$.

The first term in the r.h.s.\ of  Eq.~(\ref{eq:exact_acq_rev}) is straight-forward to evaluate; it is simply the entropy of the predictive distribution $p(\mathbf{y}|\mathcal{D},\mathbf{x})$, which is a factorizable $K$-dimensional Gaussian distribution.
Thus, we have that
\vspace{-.2cm}
\begin{align}
H(\mathbf{y}|\mathcal{D},\mathbf{x}) = \frac{K}{2} \log(2\pi e) + \sum_{i=1}^K 0.5 \log(v_k^\text{PD})\,,
\label{eq:pred_entropy}
\end{align} 
\\[-0.25cm]
where $v_k^\text{PD}$ is the predictive variance of $f_k(\cdot)$ at $\mathbf{x}$.
The difficulty comes from the evaluation of the second term in the r.h.s.\ of Eq.~(\ref{eq:exact_acq_rev}), which is intractable and must be approximated; we follow \citet{NIPS2014_5324} and approximate the expectation using a Monte Carlo estimate of the Pareto set, $\mathcal{X}^\star$ given $\mathcal{D}$.
This involves sampling several times the objective functions from their posterior distribution~$p(f_1,\ldots,f_K|\mathcal{D})$.
This step is done as in \citet{NIPS2014_5324} using random kernel features and linear  models that accurately approximate the 
samples from~$p(f_1,\ldots,f_K|\mathcal{D})$. 
In practice, we generate 10 samples from the posterior of each objective $f_k(\cdot)$.

Given the samples of the objectives, we must optimize them to obtain a sample from the Pareto set~$\mathcal{X}^\star$.
Note that unlike the true objectives, the sampled functions can be evaluated without significant cost. 
Thus, given these functions, we use a grid search with~${d \times 1,000}$ points to solve the corresponding multi-objective problem to find~$\mathcal{X}^\star$, where~$d$ is the number of dimensions.
Of course, in high dimensional problems such a grid search is expected to be sub-optimal; in that case, we use the NSGA-II evolutionary algorithm~\citep{deb2002}.
The Pareto set is then approximated using a representative subset of~$50$ points.
Given such a sample of~$\mathcal{X}^\star$, the differential entropy of~$p(\mathbf{y}|\mathcal{D},\mathbf{x}, \mathcal{X}^\star)$ is estimated using the expectation propagation algorithm \cite{Minka01}, as described in the proceeding section.

\subsection{Approximating the Conditional Predictive Distribution Using Expectation Propagation}

To approximate the entropy of the conditional predictive distribution~$p(\mathbf{y}|\mathcal{D},\mathbf{x}, \mathcal{X}^\star)$ we consider the distribution~$p(\mathcal{X}^\star|f_1,\ldots,f_K)$.
In particular,~$\mathcal{X}^\star$ is the Pareto set of~$f_1,\ldots,f_K$ iff~${\forall \mathbf{x}^\star \in \mathcal{X}^\star, 
\forall \mathbf{x}' \in \mathcal{X}, \exists\, k \in 1,\ldots,K}$ such that $f_k(\mathbf{x}^\star) \leq f_k(\mathbf{x}')$,
assuming minimization.
That is, each point within the Pareto set has to be better or equal to any other point in the domain of the functions in at least one of the objectives.
Let~$\mathbf{f}$ be the set~$\{f_1,\ldots,f_K\}$.
Informally, the conditions just described can be translated into the following un-normalized distribution for $\mathcal{X}^\star$:
\vspace{-.25cm}
\begin{align}
p(\mathcal{X}^\star|\mathbf{f}) & \propto \prod_{\mathbf{x}^\star \in \mathcal{X}^\star}
	\prod_{\mathbf{x}' \in \mathcal{X}} \left[1 - \prod_{k=1}^K\Theta
	\left(f_k(\mathbf{x}^\star) - f_k(\mathbf{x}')\right)\right]
	\nonumber \\
	& =  \prod_{\mathbf{x}^\star \in \mathcal{X}^\star}
	\prod_{\mathbf{x}' \in \mathcal{X}} \psi(\mathbf{x}',\mathbf{x}^\star)
	\label{eq:constraints}
	\,,
\end{align}
\\[-0.35cm]
where ${\psi(\mathbf{x}',\mathbf{x}^\star) = 1 - \prod_{k=1}^K\Theta \left(f_k(\mathbf{x}^\star) - f_k(\mathbf{x}')\right)}$, $\Theta(\cdot)$ is the Heaviside step function, and we have used the convention that~${\Theta(0)=1}$.
Thus, the r.h.s.\ of Eq.~(\ref{eq:constraints}) is non-zero only for a valid Pareto set.
Next, we note that in the noiseless case~$p(\mathbf{y}|\mathbf{x},\mathbf{f})=\prod_{i=1}^K \delta(y_k - f_k(\mathbf{x}))$, where~$\delta(\cdot)$ is the Dirac delta function; in the noisy case we simply replace the delta functions with Gaussians.
We can hence write the unnormalized version of~$p(\mathbf{y}|\mathcal{D},\mathbf{x}, \mathcal{X}^\star)$ as:
\vspace{-.25cm}
\begin{align}
p(\mathbf{y}|\mathcal{D},\mathbf{x}, \mathcal{X}^\star) & \propto 
	\int p(\mathbf{y}|\mathbf{x},\mathbf{f}) p(\mathcal{X}^\star|\mathbf{f}) p(\mathbf{f}|\mathcal{D}) d\mathbf{f}
\nonumber \\
& \propto \int \prod_{i=1}^K \delta(y_k - f_k(\mathbf{x})) \prod_{\mathbf{x}^\star \in \mathcal{X}^\star}
	\psi(\mathbf{x},\mathbf{x}^\star)
	\nonumber \\
        & \quad\quad \times \!\!\!\!\prod_{\mathbf{x}'  \in \mathcal{X} \setminus \{\mathbf{x}\}}\!\!\!\! 
	\psi(\mathbf{x}',\mathbf{x}^\star)\,
	p(\mathbf{f}|\mathcal{D})\,
	d \mathbf{f}
	\,,
	\label{eq:un_conditioned}
\end{align}
\\[-0.35cm]
where we have separated out the factors $\psi$ that do not depend on $\mathbf{x}$, the point in which the acquisition function $\alpha(\cdot)$ is going to be evaluated.
The approximation to the r.h.s.\ of Eq.~(\ref{eq:un_conditioned}) is obtained in two stages.
First, we approximate~$\mathcal{X}$ with the set~${\tilde{\mathcal{X}}=\{\mathbf{x}_n\}_{n=1}^N \cup \mathcal{X}^\star \cup \{\mathbf{x}\}}$, \emph{i.e.}, the union of the input locations where the objective functions have been already evaluated, the current  Pareto set and the candidate location~$\mathbf{x}$ on which~$\alpha(\cdot)$ should be evaluated.
Then, we replace each non-Gaussian factor~$\psi$ with a corresponding approximate Gaussian factor~$\tilde{\psi}$ whose  parameters are found using expectation propagation (EP) \cite{Minka01}.
That is,
\vspace{-.35cm}
\begin{align}
\psi(\mathbf{x}',\mathbf{x}^\star) & = 1 - \prod_{k=1}^K\Theta
	\left(f_k(\mathbf{x}^\star) - f_k(\mathbf{x}')\right) 
\nonumber \\
& \approx 
\tilde{\psi}(\mathbf{x}',\mathbf{x}^\star)  = \prod_{k=1}^K \tilde{\phi}_k(f_k(\mathbf{x}'), f_k(\mathbf{x}^\star))\,,
\end{align}
\\[-.35cm]
where each approximate factor~$\tilde{\phi}_k$ is an unnormalized two-dimensional Gaussian distribution.
In particular, we set
$\tilde{\phi}_k(f_k(\mathbf{x}'), f_k(\mathbf{x}^\star)) = \exp \left\{ -
	\frac{1}{2}\bm{\upsilon}_k^\text{T} \tilde{\mathbf{V}}_k \bm{\upsilon}_k+ 
	\tilde{\mathbf{m}}_k^\text{T} \bm{\upsilon}_k \right\}$,
where we have defined~${\bm{\upsilon}_k=(f_k(\mathbf{x}'), f_k(\mathbf{x}^\star))^\text{T}}$, and~$\tilde{\mathbf{V}}_k$ and~$\tilde{\mathbf{m}}_k$ are parameters to be adjusted by EP, which refines each~$\tilde{\psi}$ until convergence to enforce that it looks similar to the corresponding exact factor~$\psi$ \citep{Minka01}. 
The approximate factors~$\tilde{\psi}$ that do not depend on the candidate input~$\mathbf{x}$ are reused multiple times to evaluate the acquisition function~$\alpha(\cdot)$, and they only have to be computed once.
The~$|\mathcal{X}^\star|$ factors that depend on~$\mathbf{x}$ must be obtained relatively 
quickly to guarantee that~$\alpha(\cdot)$ is not very expensive to evaluate. 
Thus, in practice we only update those factors once using EP, i.e., they are not refined until convergence.

Once EP has been run, we approximate~$p(\mathbf{y}|\mathcal{D},\mathbf{x}, \mathcal{X}^\star)$ by the normalized Gaussian that results from replacing each exact factor~$\psi$ by the corresponding approximate~$\tilde{\psi}$. 
Note that the Gaussian distribution is closed under the product operation, and because all non-Gaussian factors in Eq.~(\ref{eq:un_conditioned}) have been replaced by Gaussians, the result is a Gaussian distribution.
That is,~${p(\mathbf{y}|\mathcal{D},\mathbf{x}, \mathcal{X}^\star) \approx \prod_{i=1}^K \mathcal{N}(f_k(\mathbf{x})|m_k^\text{CPD},v_k^\text{CPD})}$, where the parameters $m_k^\text{CPD}$ and $v_k^\text{CPD}$ can be obtained from each  $\tilde{\psi}$ and $p(f_1,\ldots,f_K|\mathcal{D})$.
If we combine this result with Eq.~(\ref{eq:pred_entropy}), we obtain an approximation to the acquisition function 
in Eq.~(\ref{eq:exact_acq_rev}) that is given by the difference in entropies before and after conditioning on the 
Pareto sets. That is,
\vspace{-.15cm}
\begin{align}
\alpha(\mathbf{x}) & \approx \sum_{k=1}^K \frac{\log v_k^\text{PD}(\mathbf{x})}{2} -  
	\frac{1}{S} \sum_{s=1}^S \frac{\log v_k^\text{CPD}(\mathbf{x}|\mathcal{X}^\star_{(s)})}{2}
\,,
\label{eq:approx}
\end{align} 
\\[-0.35cm]
where $S$ is the number of Monte Carlo samples, $\{\mathcal{X}^\star_{(s)}\}_{s=1}^S$ are the Pareto sets sampled to approximate 
the expectation in Eq.~(\ref{eq:exact_acq_rev}), and~$v_k^\text{PD}(\mathbf{x})$ and~$v_k^\text{CPD}(\mathbf{x}|\mathcal{X}^\star_{(s)})$ are respectively the variances of the predictive distribution at~$\mathbf{x}$, before and after conditioning to~$\mathcal{X}^\star_{(s)}$.
Last, in the case of noisy observations around each~$f_k(\cdot)$, we just increase the predictive variances by adding the noise variance.
The next evaluation is simply set to~${\mathbf{x}_{N+1}=\text{arg max}_{\mathbf{x} \in \mathcal{X}} \, \alpha(\mathbf{x})}$.

Note that Eq.~(\ref{eq:approx}) is the sum of~$K$ functions
\\[-.35cm]
\begin{align}
\alpha_k(\mathbf{x}) & = \frac{ \log v_k^\text{PD}(\mathbf{x})}{2} - 
\frac{1}{S}\sum_{s=1}^S \frac{\log v_k^\text{CPD}(\mathbf{x}|\mathcal{X}^\star_{(s)})}{2}
\,,
\end{align}
\\[-0.35cm]
that intuitively measure the contribution of each objective to the total acquisition.
In a decoupled evaluation setting, each~$\alpha_k(\cdot)$ can be individually maximized to  identify the location~${\mathbf{x}_k^\text{op} = {\text{arg max}}_{\mathbf{x}\in\mathcal{X}} \,\, \alpha_k(\mathbf{x})}$, on which it is expected to be most useful to evaluate each of the $K$ objectives.
The objective~$k$ with the largest individual acquisition~$\alpha_k(\mathbf{x}_k^\text{op})$ can then be chosen for evaluation in the next iteration.
This approach is expected to reduce the entropy of the posterior over the Pareto set more quickly, \emph{i.e.}, with a smaller number of evaluations of the objectives, and to lead to better results.

The total computational cost of evaluating the acquisition function~$\alpha(\mathbf{x})$ includes the cost of running EP, which is~$\mathcal{O}(Km^3)$, where~${m=N+|\mathcal{X}_{(s)}^\star|}$,~$N$ is the number of observations made and~$K$ is the number of objectives.
This is done once per each sample~$\mathcal{X}_{(s)}^\star$.
After this, we can re-use the factors that are independent of the candidate location~$\mathbf{x}$.
The cost of computing the predictive variance at each~$\mathbf{x}$ is hence~$\mathcal{O}(K|\mathcal{X}_{(s)}^\star|^3)$.
In our experiments, the size of the Pareto set sample~$\mathcal{X}_{(s)}^\star$ is 50, which means that~$m$ is a few hundred at most. 
The supplementary material contains additional details about the EP approximation to Eq.~(\ref{eq:un_conditioned}).

\section{Related Work}
\label{sec:related}

ParEGO is another method for multi-objective Bayesian optimization \cite{knowles2006}.
ParEGO transforms the multi-objective problem  into a single-objective problem using a scalarization technique: at each iteration, a vector of~$K$ weights~${\bm{\theta}=(\theta_1,\ldots,\theta_K)^\text{T}}$, with~${\theta_k \in [0,1]}$ and~${\sum_{k=1}^K \theta_k = 1}$, is sampled at random from a uniform distribution.
Given~$\bm{\theta}$, a single-objective function is built:
\vspace{-.25cm}
\begin{align}
f_{\bm{\theta}}(\mathbf{x}) = \text{max}_{k=1}^K (\theta_k f_k(\mathbf{x})) + \rho \sum_{k=1}^K \theta_k f_k(\mathbf{x})
\end{align}
\\[-.25cm]
where $\rho$ is set equal to $0.05$.
See \citep[Sec. 1.3.3]{Nakayama2009} for further details.
After step $N$ of the optimization process, and given $\bm{\theta}$, a new set of $N$ observations of $f_{\bm{\theta}}(\cdot)$ are obtained by evaluating this function in the already observed points~$\{\mathbf{x}_{n}\}_{n=1}^N$.
Then, a GP model is fit to the new data and expected improvement \cite{mockus1978application,Jones98} is used find
the location of the next evaluation~$\mathbf{x}_{N+1}$.
The cost of evaluating the acquisition function in ParEGO is $\mathcal{O}(N^3)$, where $N$ is the number of observations made.
This is the cost of fitting the GP to the new data (only done once).
Thus, ParEGO is a simple technique that leads to a fast acquisition function.
Nevertheless, it is often outperformed by more advanced approaches \citep{Ponweiser2008}.

SMSego is another technique for multi-objective Bayesian optimization \cite{Ponweiser2008}.
The first step in SMSego is to find a set of Pareto points~$\tilde{\mathcal{X}}^\star$,
\emph{e.g.}, by optimizing the posterior means of the GPs, or by finding the non-dominated observations.
Consider now an optimistic estimate of the objectives at input location~$\mathbf{x}$ given 
by~${m^\text{PD}_k(\mathbf{x}) - c \cdot v_k^\text{PD}(\mathbf{x})^{1/2}}$, where~$c$ is some constant, 
and~$m^\text{PD}_k(\mathbf{x})$ and~$v_k^\text{PD}(\mathbf{x})$ are the posterior mean and variance 
of the $k$th objective at location $\mathbf{x}$, respectively.
The acquisition value computed at a candidate location~$\mathbf{x}\in \mathcal{X}$ by SMSego is given 
by the gain in hyper-volume obtained by the corresponding optimistic estimate, after an $\epsilon$-correction has been made.
The hyper-volume is simply the volume of points in functional space above the Pareto front
(this is simply the function space values associated to the Pareto set), with respect to 
a given reference point \cite{Zitzler99}. Because the hyper-volume is maximized by the actual Pareto set, 
it is a natural measure of performance. Thus, SMSego does not reduce the problem to a single-objective. However, at 
each iteration it has to find a set of Pareto points and to fit a different GP to each one of the 
objectives. This gives a computational cost that is $\mathcal{O}(KN^3)$. Finally, evaluating the gain 
in hyper-volume at each candidate location~$\mathbf{x}$ is also more expensive than the computation of expected improvement in ParEGO.

A similar method to SMSego is the Pareto active learning (PAL) algorithm \cite{Zuluaga13}.
At iteration~$N$, PAL uses the GP prediction for each point~$\mathbf{x}\in\mathcal{X}$ to maintain an uncertainty region~$\mathcal{R}_N(\mathbf{x})$ about the objective values associated with $\mathbf{x}$.
This region is defined as the intersection of~$\mathcal{R}_{N-1}(\mathbf{x})$, \emph{i.e.}, the uncertainty region in the previous iteration, and~$\mathcal{Q}_{c}(\mathbf{x})$, defined as as the hyper-rectangle with lower-corner given by~${m^\text{PD}_k(\mathbf{x}) - c \cdot v_k^\text{PD}(\mathbf{x})^{0.5}}$, for~${k=1,\ldots,K}$, and upper-corner given by~${m^\text{PD}_k(\mathbf{x}) + c \cdot v_k^\text{PD}(\mathbf{x})^{1/2}}$, for~${k=1,\ldots,K}$, for some constant~$c$.
Given these regions, PAL classifies each point~$\mathbf{x}\in\mathcal{X}$ as Pareto-optimal, non-Pareto-optimal or uncertain.
A point is classified as Pareto-optimal if the worst value in~$\mathcal{R}_N(\mathbf{x})$ is not dominated by the best value in~$\mathcal{R}_N(\mathbf{x}')$, for any other~${\mathbf{x}'\in\mathcal{X}}$, with an~$\epsilon$ tolerance.
A point is classified as non-Pareto-optimal if the best value in~$\mathcal{R}_N(\mathbf{x})$ is dominated by the worst value in~$\mathcal{R}_N(\mathbf{x}')$ for any other~${\mathbf{x}'\in\mathcal{X}}$, with an~$\epsilon$ tolerance.
All other points remain uncertain.
After the classification, PAL chooses the uncertain point~$\mathbf{x}$ with the largest uncertainty 
region~$\mathcal{R}_N(\mathbf{x})$.  The total computational cost of PAL is hence similar to that of SMSego.

The expected hyper-volume improvement (EHI) \cite{Emm08} 
is a natural extension of expected improvement to the multi-objective setting \cite{mockus1978application,Jones98}.
Given the predictive distribution of the GPs at a candidate input location~$\mathbf{x}$, the acquisition is the 
expected increment of the hyper-volume of a candidate Pareto set~$\tilde{\mathcal{X}}^\star$.
Thus, EHI also needs to find a Pareto set~$\tilde{\mathcal{X}}^\star$. This set can be obtained as in 
SMSego.  A difficulty is, however, that computing the expected increment of the hyper-volume is very expensive.
For this, the output space is divided in a series of cells, and the probability of improvement is simply 
obtained as the probability that the observation made at~$\mathbf{x}$ lies in a non-dominated cell.
This involves a sum across all non-dominated cells, whose number grows exponentially with the number of 
objectives~$K$. In particular, the total number of cells is~$(|\tilde{\mathcal{X}}^\star| + 1)^K$.
Thus, although some methods have been suggested to speed-up its calculation, \emph{e.g.}, \cite{Hupkens2014}, EHI is only 
feasible for a 2 or 3 objectives at most.

Sequential uncertainty reduction (SUR) is another method proposed for multi-objective Bayesian optimization \cite{Picheny2015}.
The working principle of SUR is similar to that of EHI. However, SUR considers the probability of improving 
the hyper-volume in the whole domain of the objectives~$\mathcal{X}$.
Thus, SUR also needs to find a set of Pareto points~$\tilde{\mathcal{X}}^\star$. These can be obtained as in SMSego.
The acquisition computed by SUR is simply the expected decrease 
in the area under the probability of improving the hyper-volume, after evaluating the objectives 
at a new candidate location~$\mathbf{x}$. The SUR acquisition is computed also by dividing the output space 
in a total of~$(|\tilde{\mathcal{X}}^\star| + 1)^K$ cells, and the area under the probability of improvement is 
obtained using a Sobol sequence as the integration points. Although some grouping of the cells has been suggested 
\cite{Picheny2015}, SUR is an extremely expensive criterion that is only feasible for 2 or 3 objectives at most. 

The proposed approach, {\small PESMO}, differs from the methods described in this section in that 1)~it does not transform the multi-objective problem into a single-objective, 2)~the acquisition function of {\small PESMO} can be decomposed as the sum of~$K$ individual acquisition functions, and this allows for decoupled evaluations, and 3)~the computational cost of {\small PESMO} is linear in the total number of objectives $K$.

\section{Experiments}
\label{sec:experiments}

We compare {\small PESMO} with the other strategies for multi-objective optimization described 
in Section \ref{sec:related}: ParEGO, SMSego, EHI and SUR.
We do not  compare results with PAL because it is expected to give similar 
results to those of SMSego, as both methods are based on a lower confidence bound.
We have coded  all these  methods in the software for Bayesian optimization Spearmint. 
In all GP models we use a Mat\'ern covariance function, and all hyper-parameters (noise, length-scales and amplitude) 
are approximately sampled from their posterior distribution (we generate 10 samples from this distribution).
The acquisition function of each method is averaged over these samples.
In ParEGO we consider a different scalarization (\emph{i.e.}, a different value 
of~$\bm{\theta}$) for each sample of the hyper-parameters.
In SMSego, EHI and SUR, for each hyper-parameter sample we consider 
a different Pareto set~$\tilde{\mathcal{X}}^\star$, obtained by optimizing the posterior means of the GPs.
The resulting Pareto set is extended by including all non-dominated observations.
Finally, at iteration~$N$, each method gives a recommendation in the form of a Pareto set 
obtained by optimizing the posterior means of the GPs (we average the posterior means over the hyper-parameter samples).
The acquisition function of each method is maximized using L-BFGS.
A grid of size $1,000$ is used to find a good starting point for the optimization process.
The gradients of the acquisition function are approximated by differences (except in ParEGO). 

\subsection{Accuracy of the {\small PESMO} Approximation}

\begin{figure}[htb]
\vspace{-.15cm}
\begin{center}
\begin{tabular}{cc}
\includegraphics[width = 0.475 \textwidth]{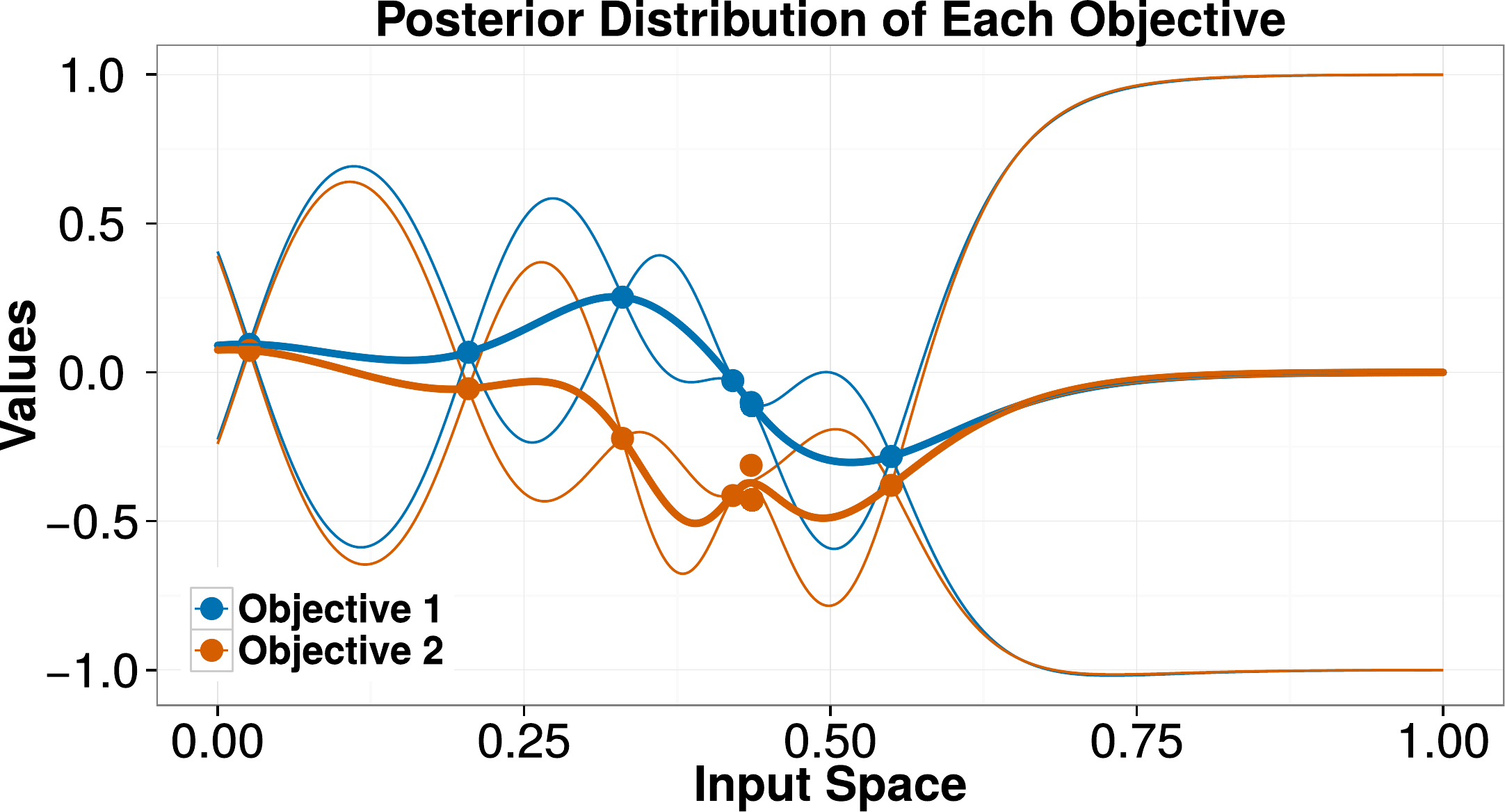}  \\
\includegraphics[width = 0.475 \textwidth]{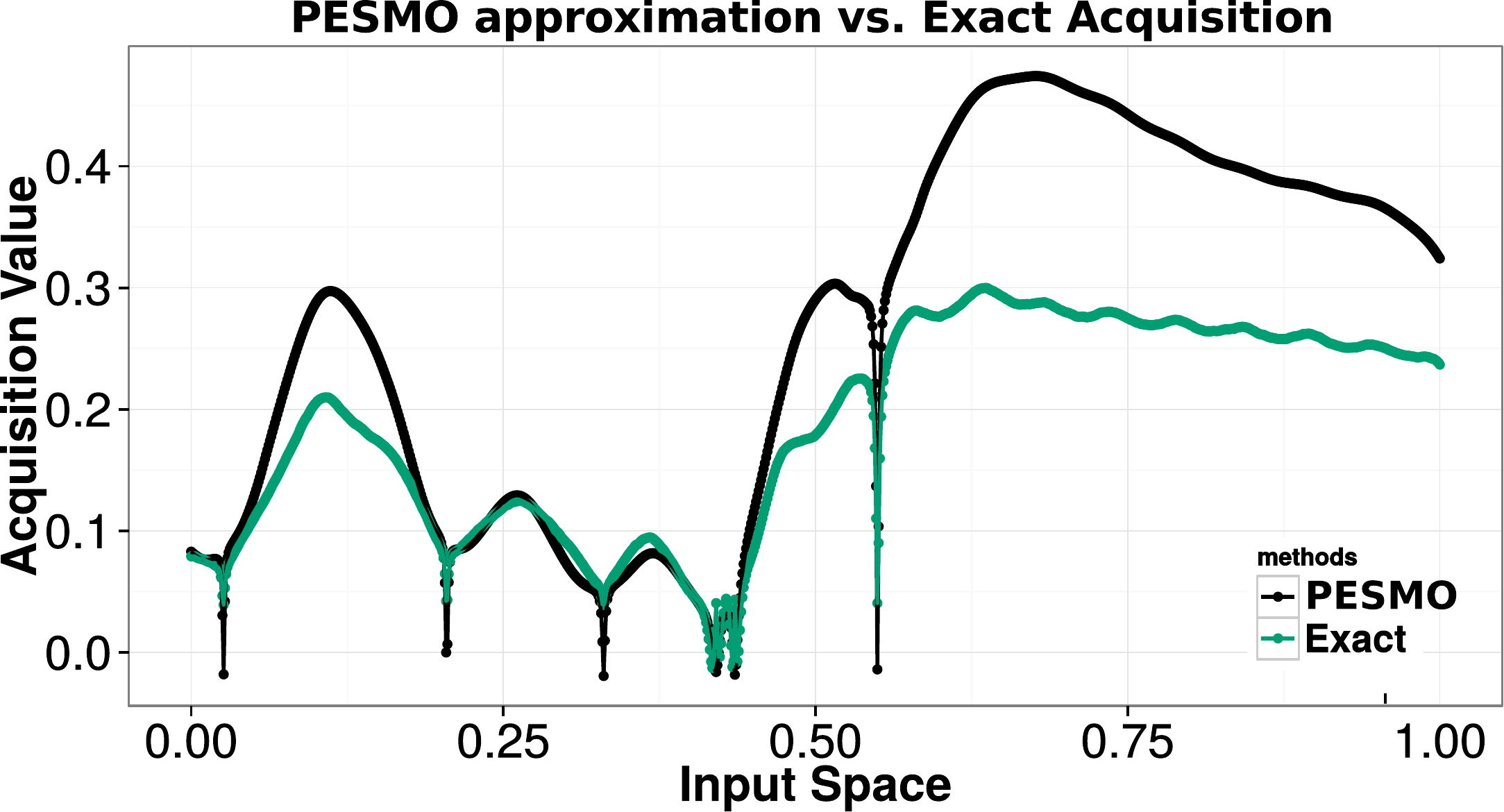}   
\end{tabular}
\end{center}
\vspace{-.35cm}
\caption{{\small (top) Observations of each objective and posterior mean and standard deviations of each 
GP model. (bottom) Estimates of the acquisition function (\ref{eq:exact_acq_rev}) by {\small PESMO}, and by
a Monte Carlo method combined with a non-parametric estimator of the entropy (Exact), which is expected to be more accurate. 
Best seen in color.}}
\label{fig:acq}
\vspace{-.35cm}
\end{figure}

One question to be experimentally addressed is whether the proposed approximations are sufficiently accurate for effective identification of the Pareto set.
We compare in a one-dimensional problem with two objectives the acquisition function computed by {\small PESMO} with a more accurate estimate obtained via expensive Monte Carlo sampling and a non-parametric estimator of the entropy \cite{singh2003nearest}.
Figure \ref{fig:acq} (top) shows at a given step the observed data and the posterior mean and the standard deviation of each of the two objectives.
The figure on the bottom shows the corresponding acquisition function computed by {\small PESMO} and by the Monte Carlo method (Exact).
We observe that both functions look very similar, including the location of the global maximizer.
This indicates that Eq.~(\ref{eq:approx}), obtained by expectation propagation, is potentially a good approximation of Eq.~(\ref{eq:exact_acq_rev}), the exact acquisition function. The supplementary material has extra results that show that each of the individual acquisition functions 
computed by {\small PESMO}, \emph{i.e.},~$\alpha_k(\cdot)$, for~${k=1,2}$, are also accurate.

\subsection{Experiments with Synthetic Objectives} 

\begin{figure*}[htb]
\begin{center}
\begin{tabular}{cc}
\includegraphics[width = 0.475 \textwidth]{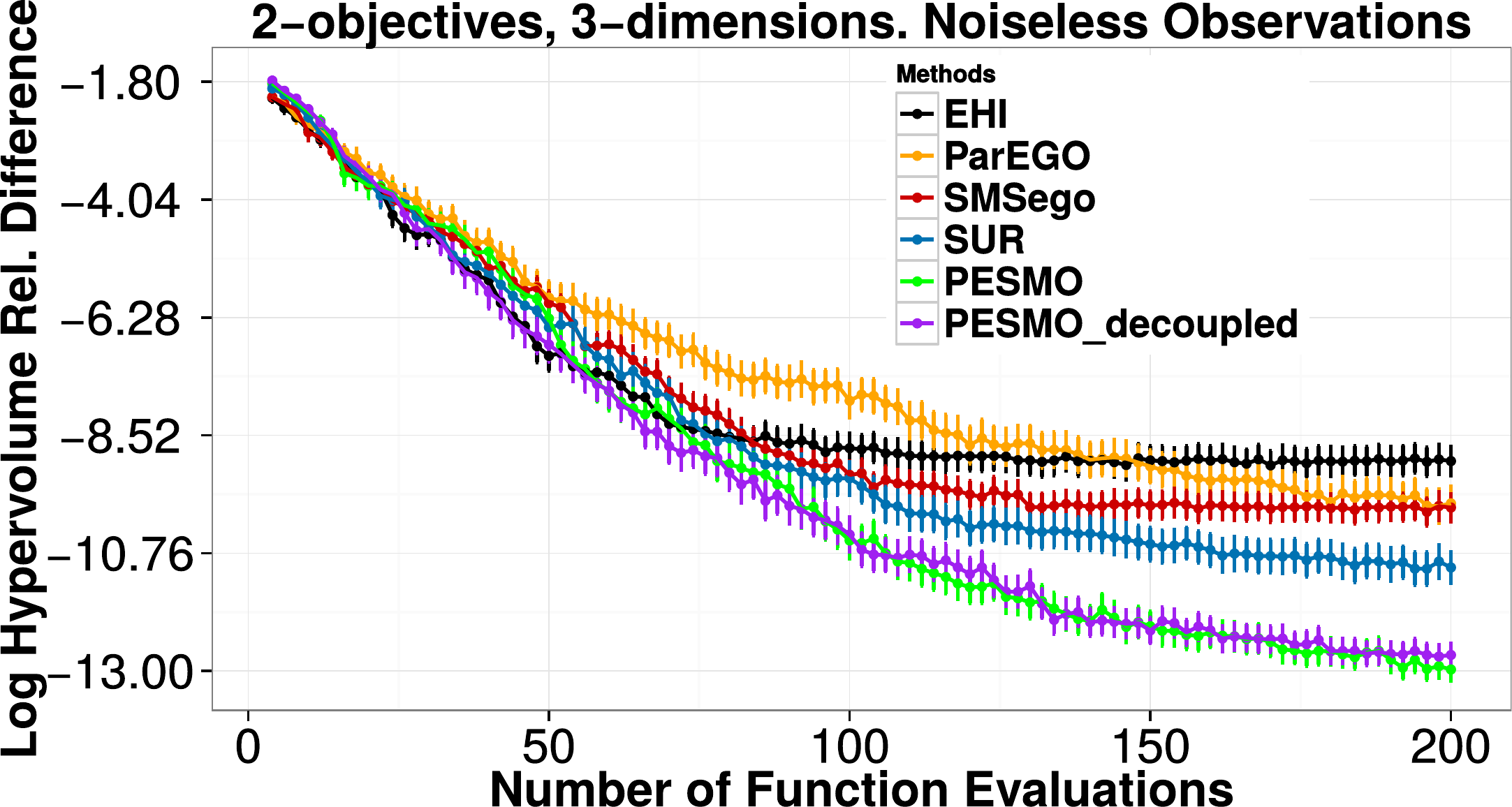}  &
\includegraphics[width = 0.475 \textwidth]{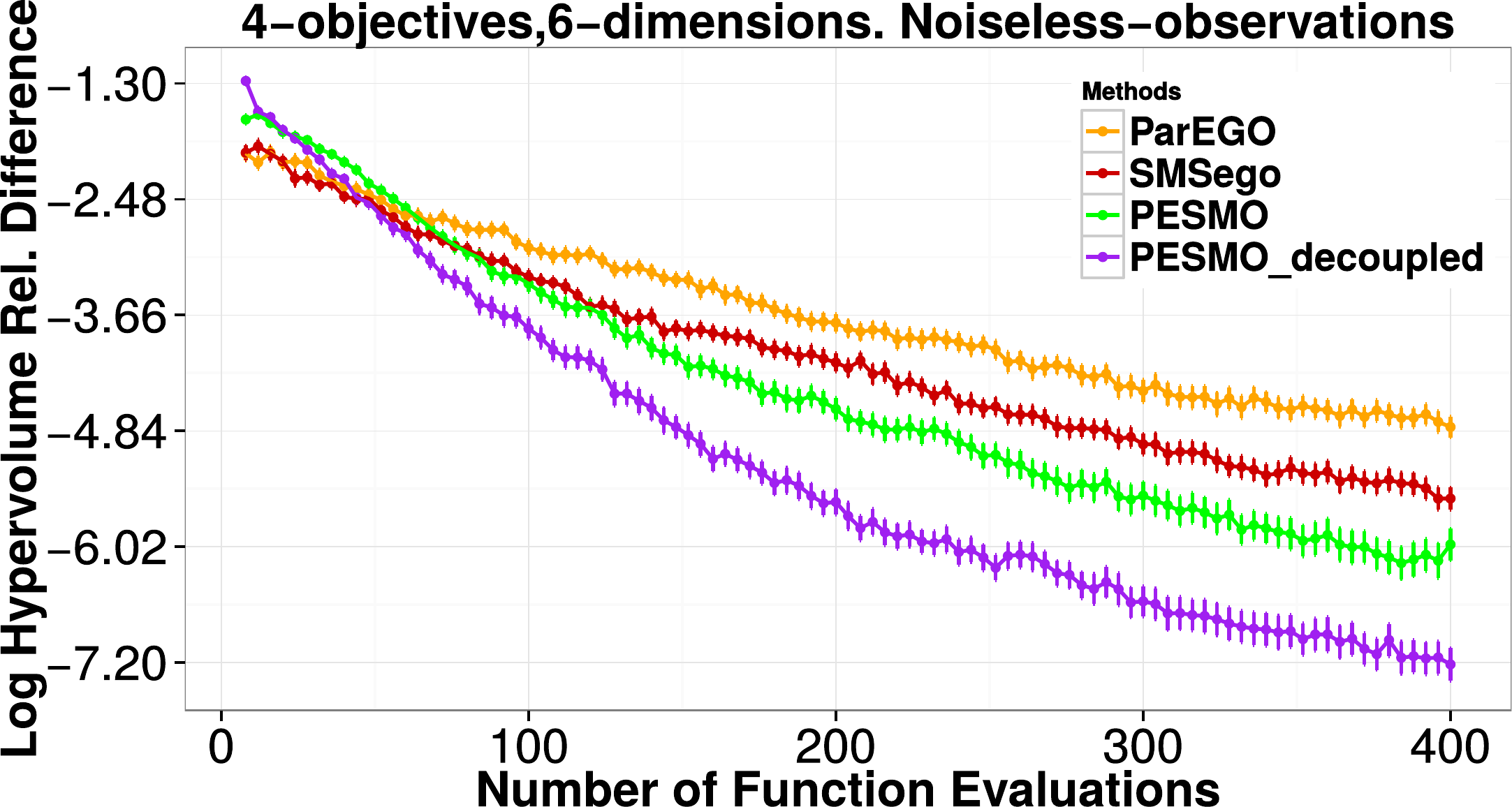}  \\
\includegraphics[width = 0.475 \textwidth]{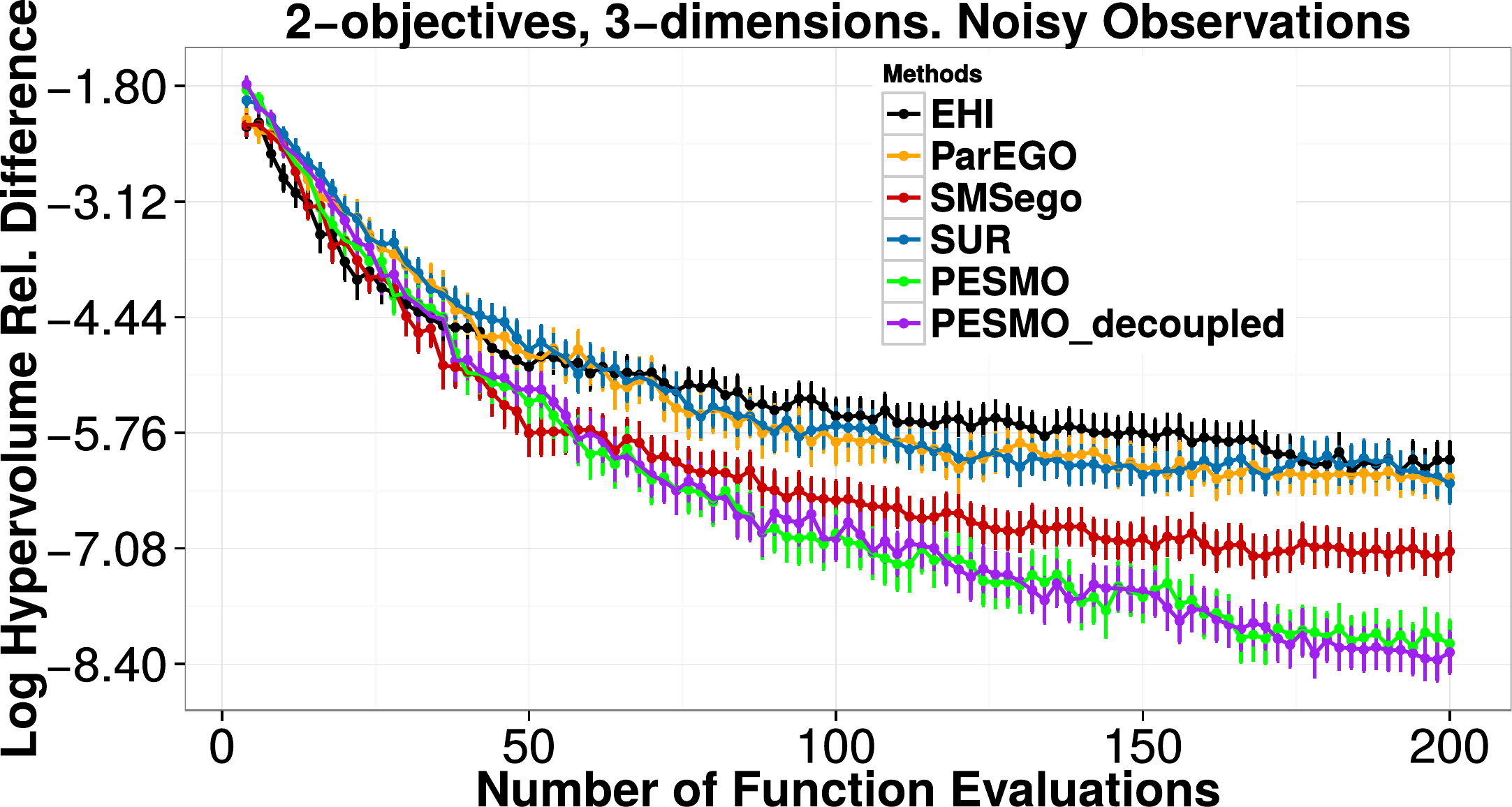}  &
\includegraphics[width = 0.475 \textwidth]{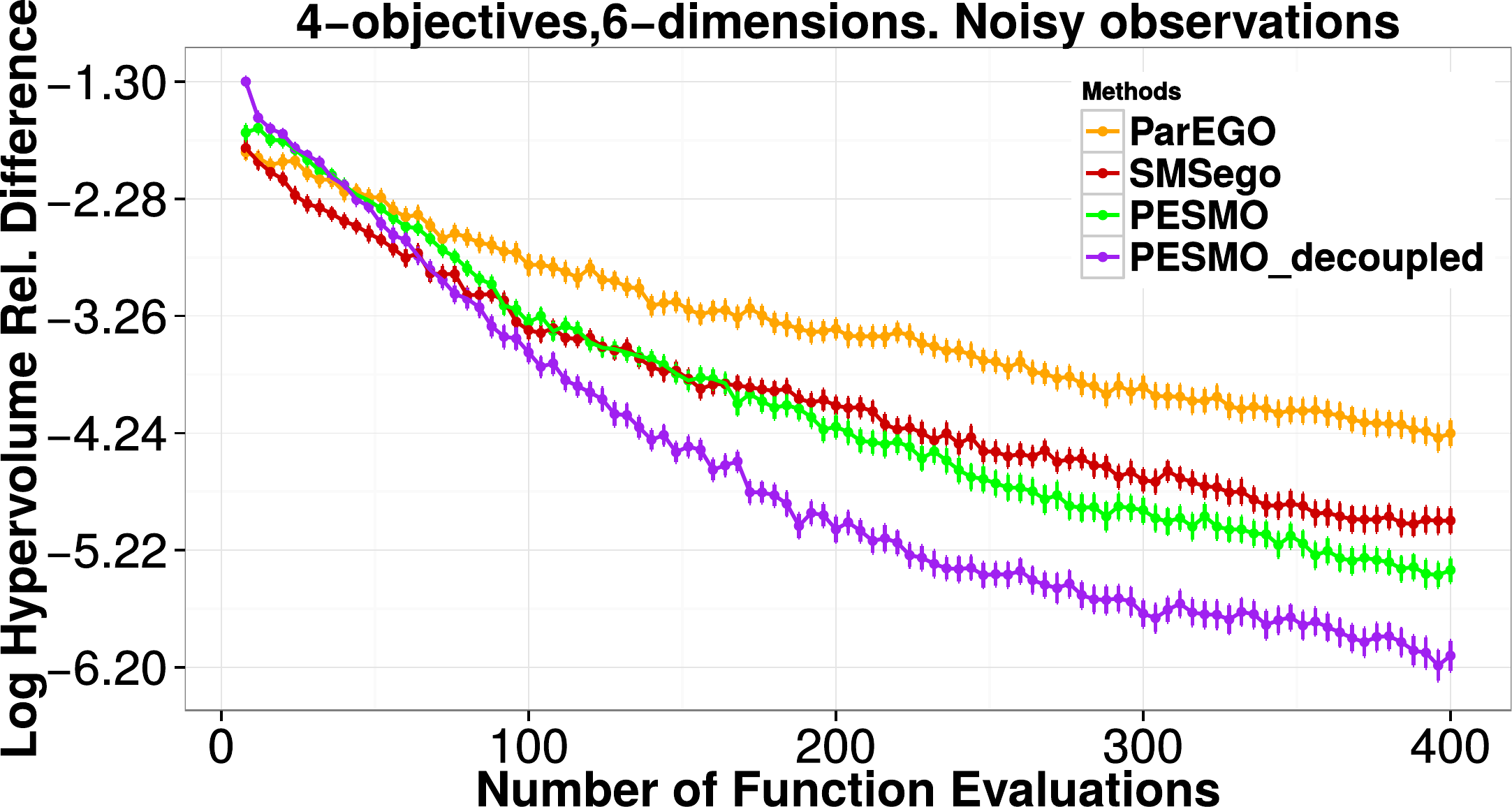}   
\end{tabular}
\end{center}
\vspace{-.35cm}
\caption{{\small (left-column) Average log relative difference between the hyper-volume of the recommendation and 
the maximum hyper-volume for each number of evaluations made. We consider noiseless (top) and noisy observations (bottom). 
The problem considered has 2 objectives and 3 dimensions. (right-column) Similar results for a problem with 
4 objectives and 6 dimensions. We do not compare results with EHI and SUR because they are infeasible due to
their exponential cost with the number of objectives. Best seen in color.}}
\label{fig:hyper_volumes}
\vspace{-.35cm}
\end{figure*}

To initially compare PEMS with other approaches, we consider a 3-dimensional problem with 2 objectives obtained by sampling the functions from the corresponding GP prior.
We generate 100 of these problems and report the average performance of each method, when considering noiseless observations and when the observations are contaminated with Gaussian noise with standard deviation equal to $0.1$.
The performance metric employed is the hyper-volume indicator, which is maximized by the actual Pareto set \cite{Zitzler99}. 
More precisely, at each iteration we report the logarithm of the relative difference between the hyper-volume of the actual Pareto set, which is obtained by optimizing the actual objectives, and the hyper-volume of the recommendation, which is obtained by optimizing the posterior means of the GPs.
Figure \ref{fig:hyper_volumes} (left-column) shows, as a function of the evaluations made, the average performance of each method with the corresponding error bars. {\small PESMO} obtains the best results, and when executed in a decoupled scenario slight improvements are observed, although only in the case of noisy observations.

Table \ref{tab:times} shows the average time in seconds required by each method to determine the next evaluation.
The fastest method is ParEGO followed by SMSego and {\small PESMO}. 
The decoupled version of {\small PESMO},~$\text{{\small PESMO}}_\text{dec}$, takes more time because it has to optimize $\alpha_1(\cdot)$ and $\alpha_2(\cdot)$.
The slowest methods are EHI and SUR; most of their cost is in the last iterations, in which the Pareto set size, 
$|\tilde{\mathcal{X}}^\star|$, is large due to non-dominated observations.
The cost of evaluating the acquisition function in EHI and SUR is~$\mathcal{O}((|\tilde{\mathcal{X}}^\star|+1)^K)$, leading to expensive 
optimization via L-BFGS. In {\small PESMO} the cost of evaluating~$\alpha(\cdot)$ is~$\mathcal{O}(K|\mathcal{X}_{(s)}^\star|^3)$ 
because~$K$ linear systems are solved. These computations are faster because they are performed using the open-BLAS library, 
which is optimized for each processor. The acquisition function of EHI and SUR does not involve solving linear systems and hence 
these methods cannot use open-BLAS. Note that we also keep fixed~$|\mathcal{X}_{(s)}^\star|=50$ in {\small PESMO}.

\vspace{-.25cm}

\begin{table}[htb]
\vspace{-.25cm}
\begin{center}
\caption{Avg. time in seconds doing calculations per iteration.}
\label{tab:times}
{\small
\begin{tabular}{@{\hspace{0mm}}r@{$\pm$}l@{\hspace{1mm}}r@{$\pm$}l@{\hspace{1mm}}
	r@{$\pm$}l@{\hspace{1mm}}r@{$\pm$}l@{\hspace{2mm}}r@{$\pm$}l@{\hspace{2mm}}r@{$\pm$}l@{\hspace{0mm}}}
\hline
\multicolumn{2}{@{\hspace{1mm}}c@{\hspace{1mm}}}{\bf PESMO} &
\multicolumn{2}{@{\hspace{1mm}}c@{\hspace{1mm}}}{\bf $\text{PESMO}_\text{dec}$} &
\multicolumn{2}{@{\hspace{0mm}}c@{\hspace{0mm}}}{\bf ParEGO} &
\multicolumn{2}{@{\hspace{1mm}}c@{\hspace{1mm}}}{\bf SMSego} &
\multicolumn{2}{@{\hspace{2mm}}c@{\hspace{2mm}}}{\bf EHI} &
\multicolumn{2}{@{\hspace{2mm}}c@{\hspace{2mm}}}{\bf SUR} \\
\hline 
33 & 1.0 & \hspace{1mm} 52 & 2.5 & \hspace{2mm} 11 & 0.2 & \hspace{2mm}16 & 1.3 & 405 & 115 & 623 & 59 \\
\hline
\end{tabular}
}
\end{center}
\end{table}

\begin{figure}[htb!]
\vspace{-.15cm}
\begin{center}
\begin{tabular}{cccc}
\includegraphics[width = 0.1 \textwidth]{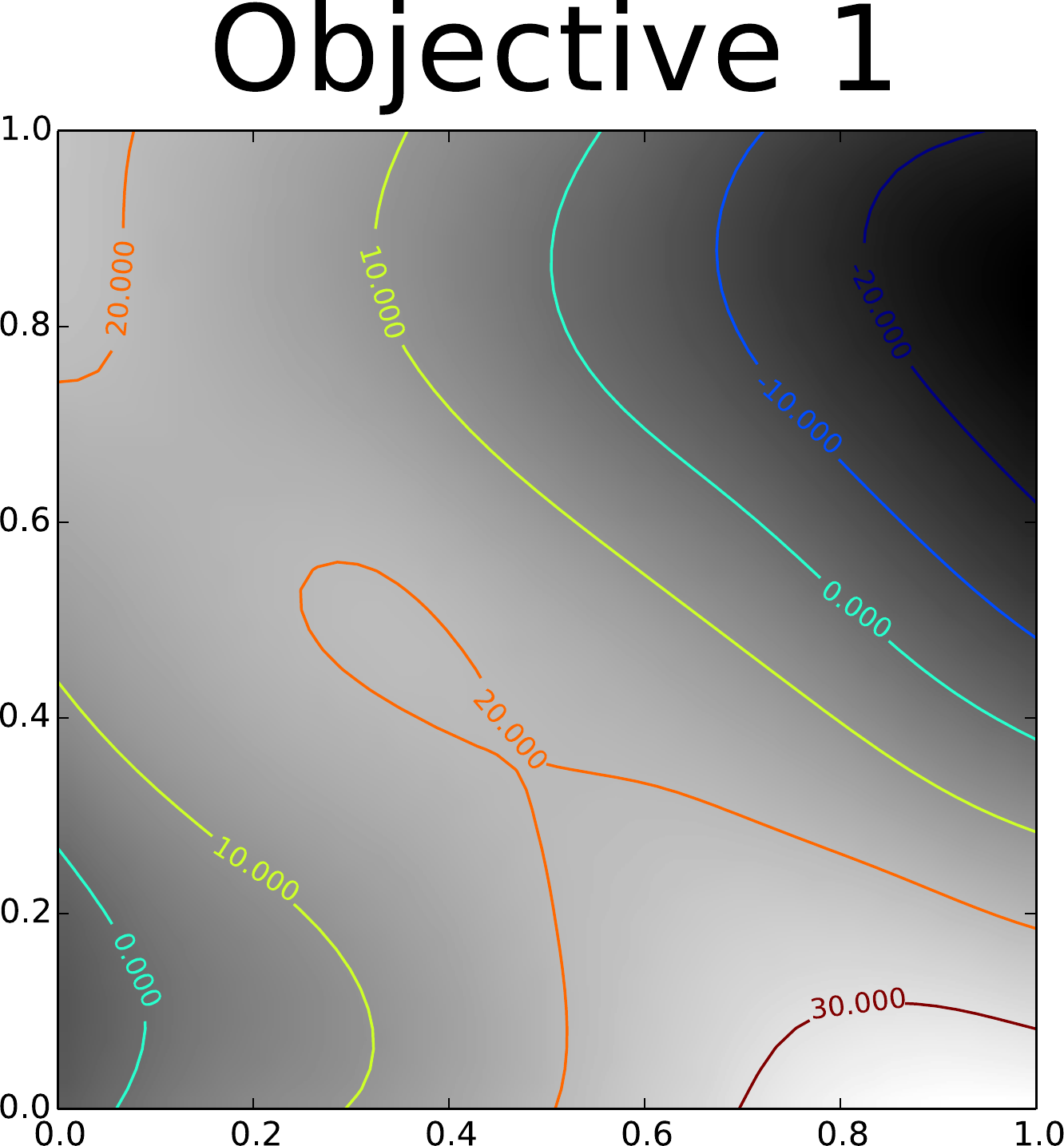}  &
\includegraphics[width = 0.1 \textwidth]{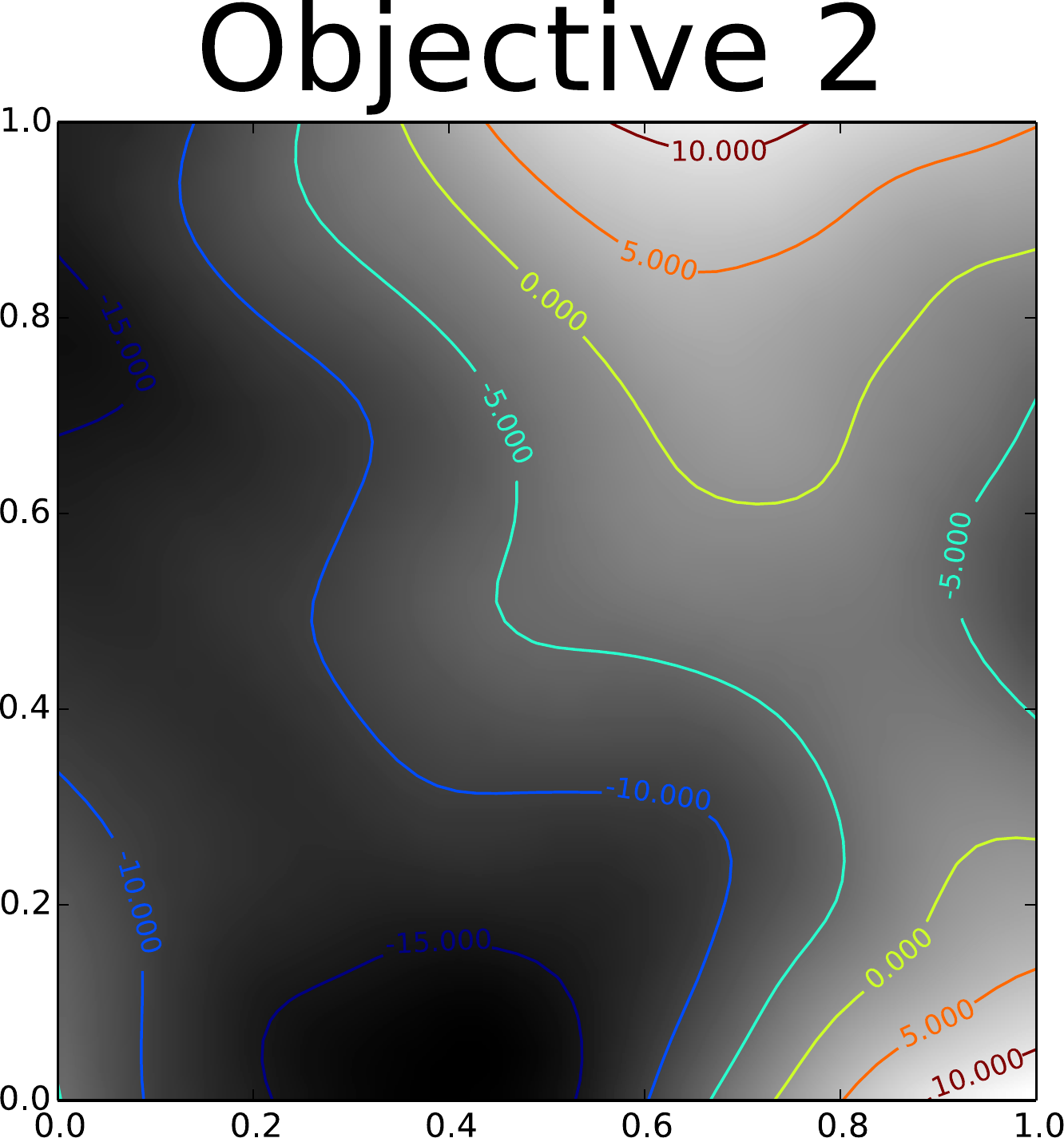}  &
\includegraphics[width = 0.1 \textwidth]{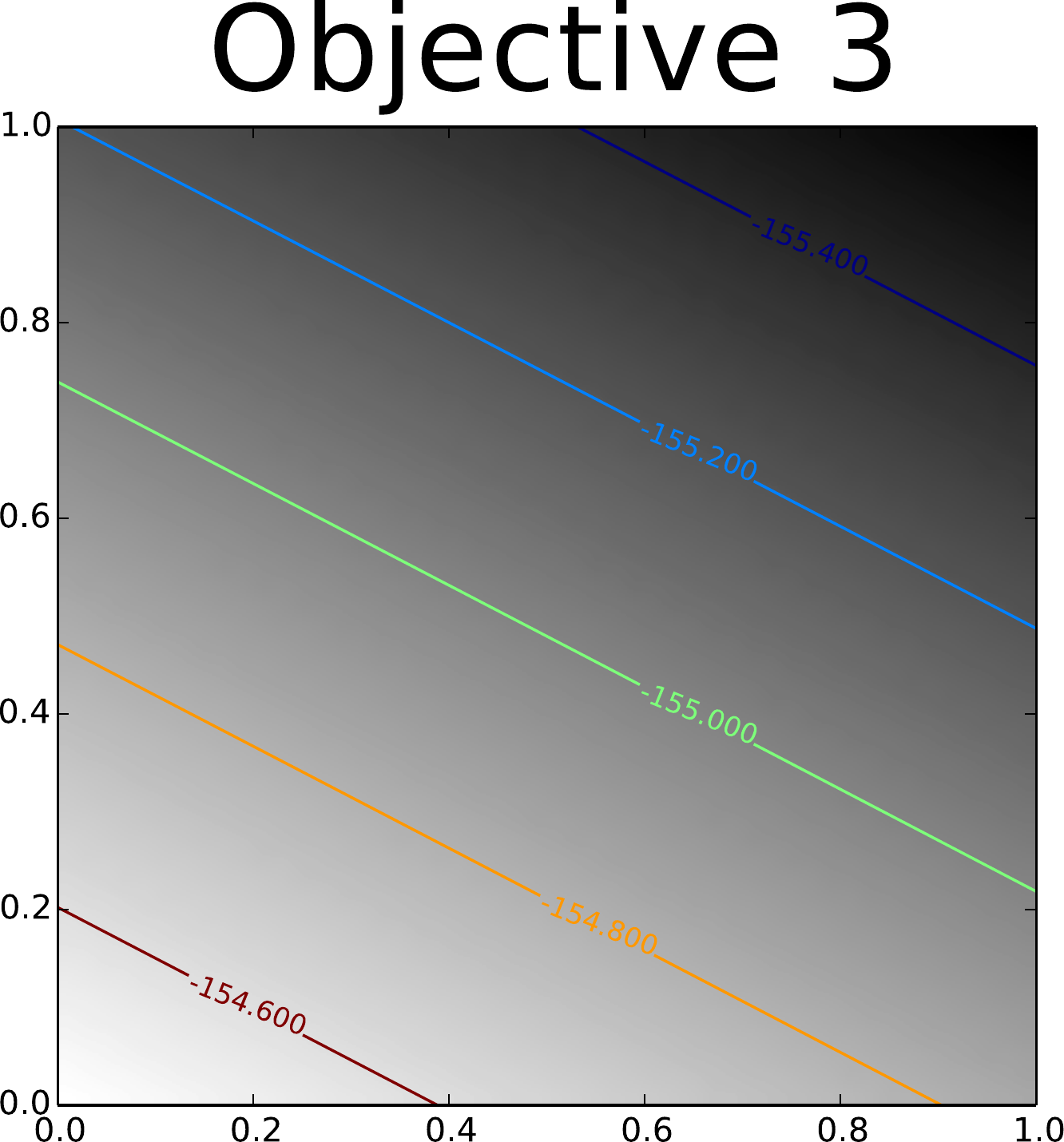}  &
\includegraphics[width = 0.1 \textwidth]{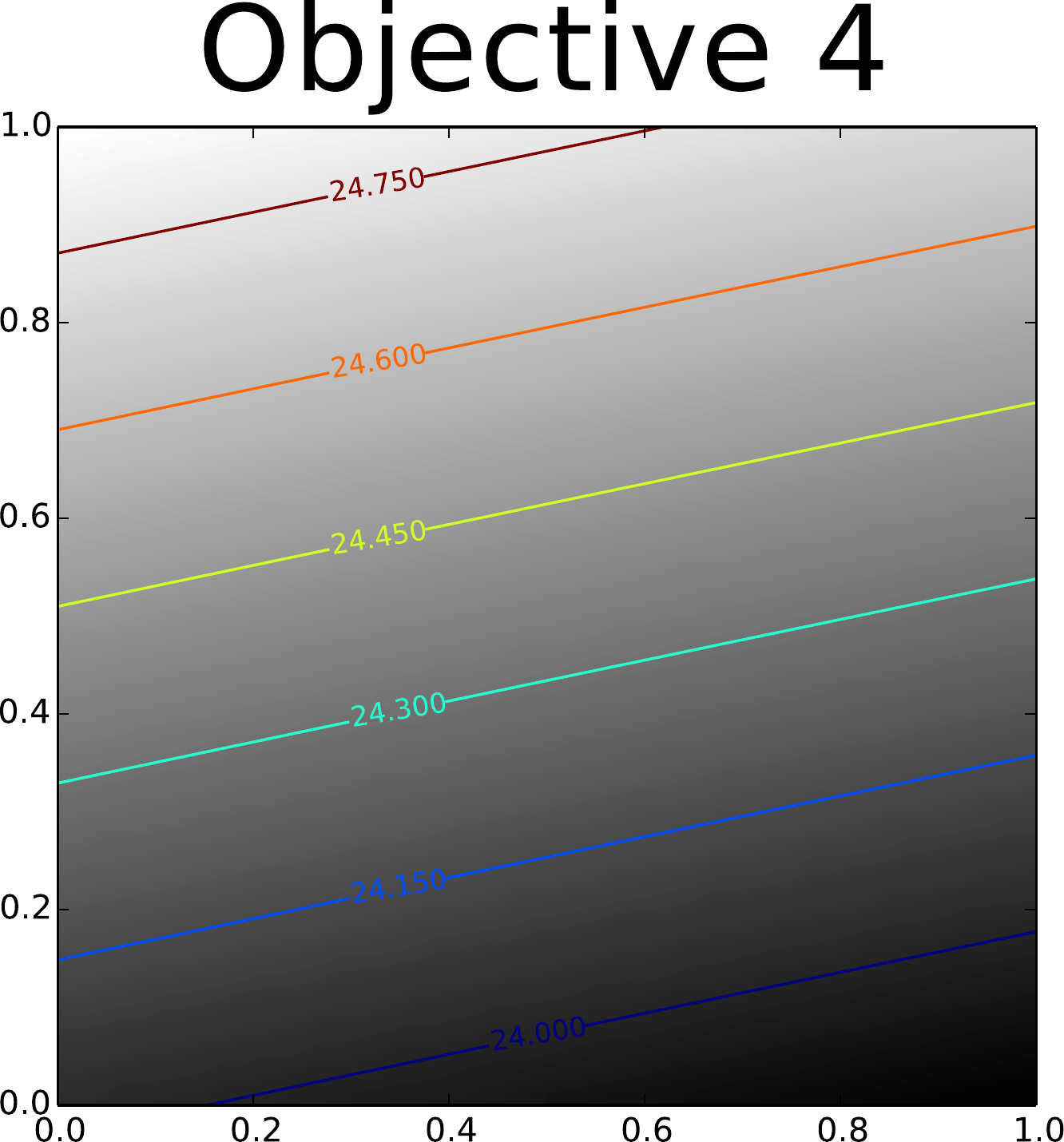}  \\
\multicolumn{4}{c}{
\includegraphics[width = 0.47 \textwidth]{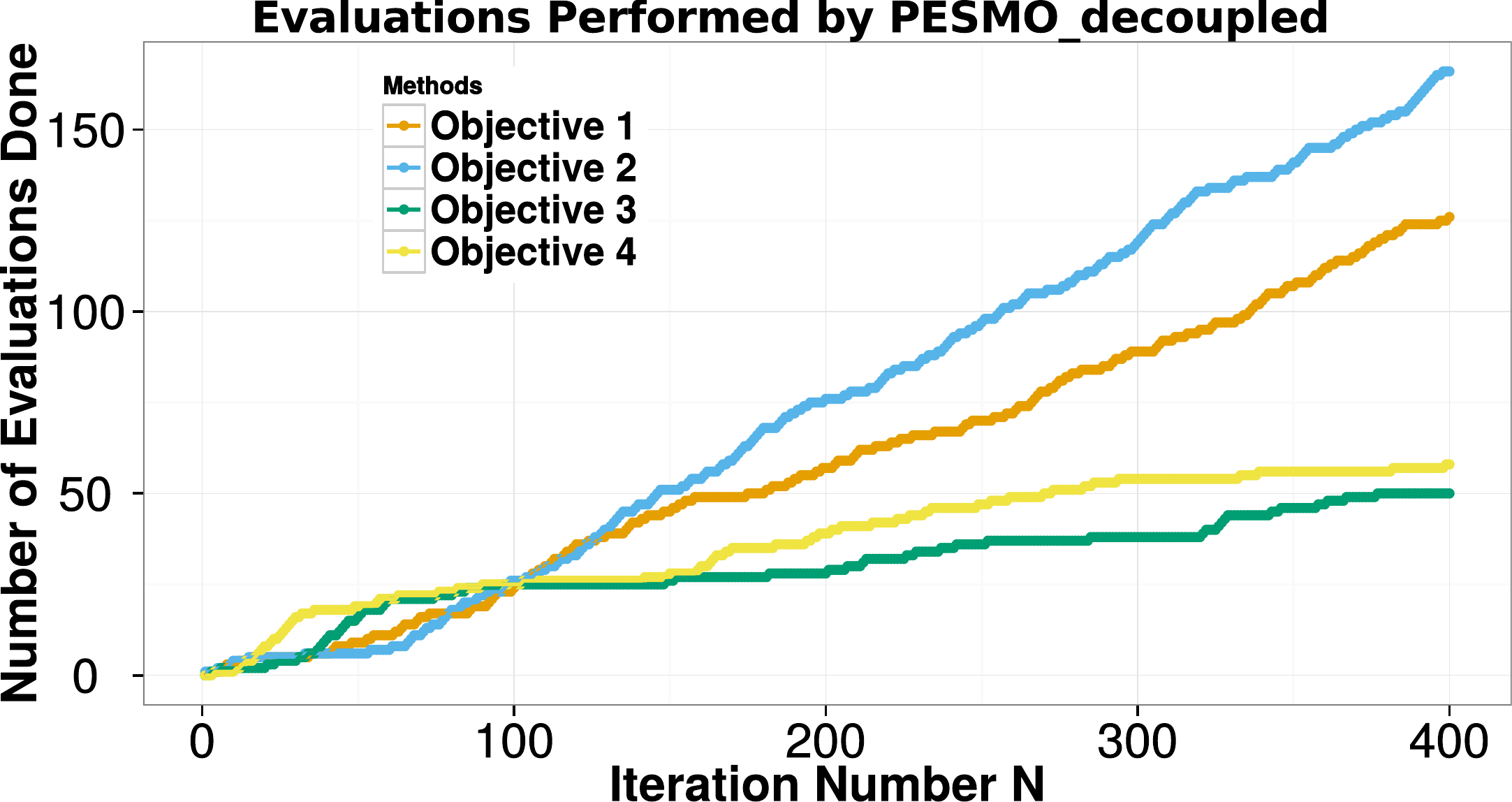}   
}
\end{tabular}
\end{center}
\vspace{-.35cm}
\caption{{\small (top) Contour curves of 4 illustrative objectives on 6 dimensions obtained by changing the first 
two dimensions in input space while keeping the other 4 fixed to zero. The first 2 objectives are non-linear
while the 2 last objectives are linear. (bottom) Number of evaluations of each objective done by $\text{{\small PESMO}}_\text{decoupled}$
as a function of the iterations performed $N$. Best seen in color.
}}
\label{fig:objectives}
\vspace{-.25cm}
\end{figure}

We have carried out additional synthetic experiments with 4 objectives on a 6-dimensional input space.
In this case, EHI and SUR become infeasible, so we do not compare results with them.
Again, we sample the objectives from the GP prior.
Figure \ref{fig:hyper_volumes} (right-column) shows, as a function of the evaluations made, the average performance of each method.
The best method is {\small PESMO}, and in this case, the decoupled evaluation performs significantly better.
This improvement is because in the decoupled setting, {\small PESMO} identifies the most difficult  objectives and evaluates them more times.
In particular, because there are 4 objectives it is likely that some objectives are more difficult than others just by chance.
Figure \ref{fig:objectives} illustrates this behavior for a representative case in which the first two objectives are non-linear (difficult) and the last two objectives are linear (easy).
We note that the decoupled version of $\text{{\small PESMO}}$ evaluates the first two objectives almost three times more.

\subsection{Finding a Fast and Accurate Neural Network}

We consider the MNIST dataset \citep{lecun98gradientbased} and evaluate each method on  the task of finding a neural network with low prediction error and small prediction time.
These are conflicting objectives because reducing the prediction error will involve larger networks which will take longer at test time.
We consider feed-forward networks with ReLus at the hidden layers and a soft-max  output layer.
The networks are coded in Keras ({\small \url{http://github.com/fchollet/keras}}) and they are trained using Adam \citep{kingma2014} with a minibatch size of $4,000$ instances during $150$ epochs.
The adjustable parameters are: The number of hidden units per layer (between $50$ and $300$), the number of layers (between $1$ and $3$), the learning rate, the  amount of dropout, and the level of $\ell_1$ and $\ell_2$ regularization. 
The prediction error is measured on a set of $10,000$ instances extracted from the training set.
The rest of the training data, \emph{i.e.}, $50,000$ instances, is used for  training.
We consider a logit transformation of the prediction error because the error rates are very small.
The prediction time is measured as the average time required for doing $10,000$ predictions.
We compute the logarithm of the ratio between the prediction time of the network and the prediction time of the fastest network, (\emph{i.e.}, a single hidden layer and $50$ units).
When measuring the prediction time we do not train the network and consider random weights (in Spearmint the time objective is also set to ignore irrelevant parameters).
Thus, the problem is suited for a decoupled evaluation because both objectives can be evaluated separately.
We run each method for a total of $200$ evaluations of the objectives and report results after $100$ and $200$ evaluations.
Because there is no ground truth and the objectives are noisy, we re-evaluate 3 times the values associated with the recommendations made by each method (in the form of a Pareto set) and average the results.
Then, we compute the Pareto front (\emph{i.e.}, the function space values of the Pareto  set) and its hyper-volume.
We repeat these experiments $50$ times and report the average results across repetitions.

\begin{figure*}[hbt]
\begin{center}
\begin{tabular}{cc}
\includegraphics[width = 0.47 \textwidth]{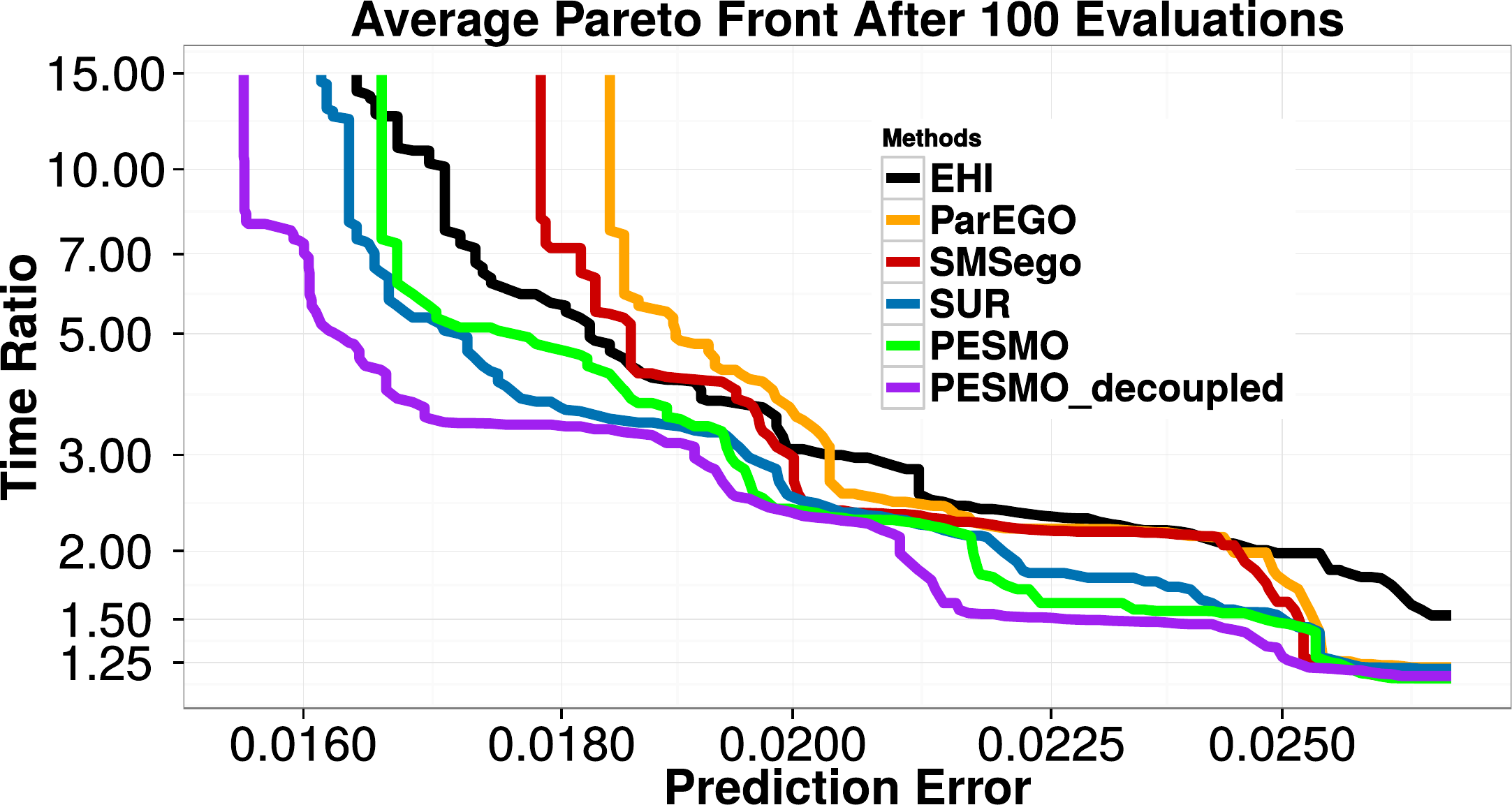}  &
\includegraphics[width = 0.47 \textwidth]{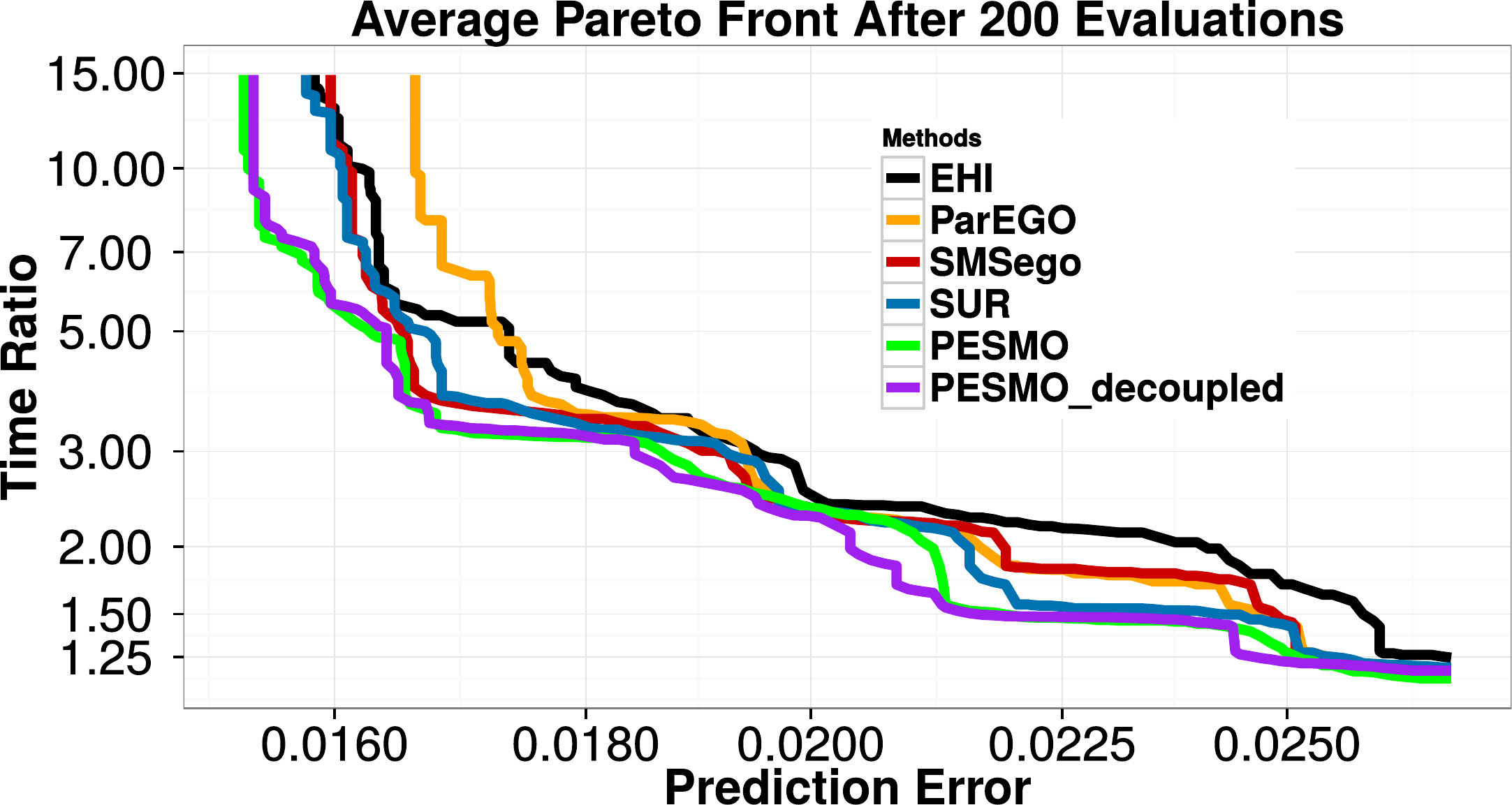}  
\end{tabular}
\end{center}
\vspace{-.4cm}
\caption{{\small Avg. Pareto fronts obtained by each method after 
100 (left) and 200 (right) evaluations of the objectives. Best seen in color. 
}}
\label{fig:frontiers}
\vspace{-.4cm}
\end{figure*}

Table \ref{tab:nnet} shows the hyper-volumes obtained in the experiments (the higher, the better). 
The best results, after $100$ evaluations of the objectives, correspond to the decoupled 
version of {\small PESMO}, followed by SUR and by the coupled version.  
When $200$ evaluations are done, the best method is {\small PESMO} in either setting, \emph{i.e.}, coupled or 
decoupled.  After {\small PESMO}, SUR gives the best results, followed by SMSego and EHI. ParEGO is the worst 
performing method in either setting. In summary, {\small PESMO} gives the best overall results, and its decoupled version 
performs much better than the other methods when the number of evaluations is small. 

\begin{table}[htb]
\vspace{-.35cm}
\caption{{\small Avg. hyper-volume after $100$ and $200$ evaluations.}}
\label{tab:nnet}
{\footnotesize
\begin{center}
\begin{tabular}{@{\hspace{0mm}}c@{\hspace{0.5mm}}r@{$\pm$}l@{\hspace{0.5mm}}r@{$\pm$}l@{\hspace{0.5mm}}
	r@{$\pm$}l@{\hspace{0.5mm}}r@{$\pm$}l@{\hspace{0.5mm}}r@{$\pm$}l@{\hspace{0.5mm}}r@{$\pm$}l@{\hspace{0mm}}}
\hline
{\bf \# Eval.} & 
\multicolumn{2}{@{\hspace{0.5mm}}c@{\hspace{0.5mm}}}{\bf PESMO} &
\multicolumn{2}{@{\hspace{0.5mm}}c@{\hspace{0.5mm}}}{\bf $\text{PESMO}_\text{dec}$} &
\multicolumn{2}{@{\hspace{0mm}}c@{\hspace{0mm}}}{\bf ParEGO} &
\multicolumn{2}{@{\hspace{0.5mm}}c@{\hspace{0.5mm}}}{\bf SMSego} &
\multicolumn{2}{@{\hspace{0.5mm}}c@{\hspace{0.5mm}}}{\bf EHI} &
\multicolumn{2}{@{\hspace{0.5mm}}c@{\hspace{0.5mm}}}{\bf SUR} \\
\hline
100 & 66.2 & .2 & \hspace{1mm}{\bf 67.6} & {\bf .1} & 62.9 & 1.2 & 65.0 & .3 & 64.0 & .9 & 66.6 & .2 \\
\hline
200 & {\bf 67.8} & {\bf .1} &\hspace{1mm} {\bf 67.8} & {\bf .1} & 66.1 & .2 & 67.1 & .2 & 66.6 & .2 & 67.2 & .1 \\
\hline
\end{tabular}
\end{center}
}
\vspace{-.35cm}
\end{table}

Figure \ref{fig:frontiers} shows the average Pareto front obtained by each method after $100$ and $200$ evaluations of the objectives.
The results displayed are consistent with the ones in Table \ref{tab:nnet}.
In particular, {\small PESMO} is able to find networks that are faster than the ones found by the other methods, for a similar prediction error on the validation set.
This is especially the case of {\small PESMO} when executed in a decoupled setting, after doing only $100$ evaluations of the objectives.
We also note that {\small PESMO} finds  the most accurate networks, with almost $1.5\%$ of prediction error in the validation set.

\begin{figure}[htb]
\begin{center}
\includegraphics[width = 0.47 \textwidth]{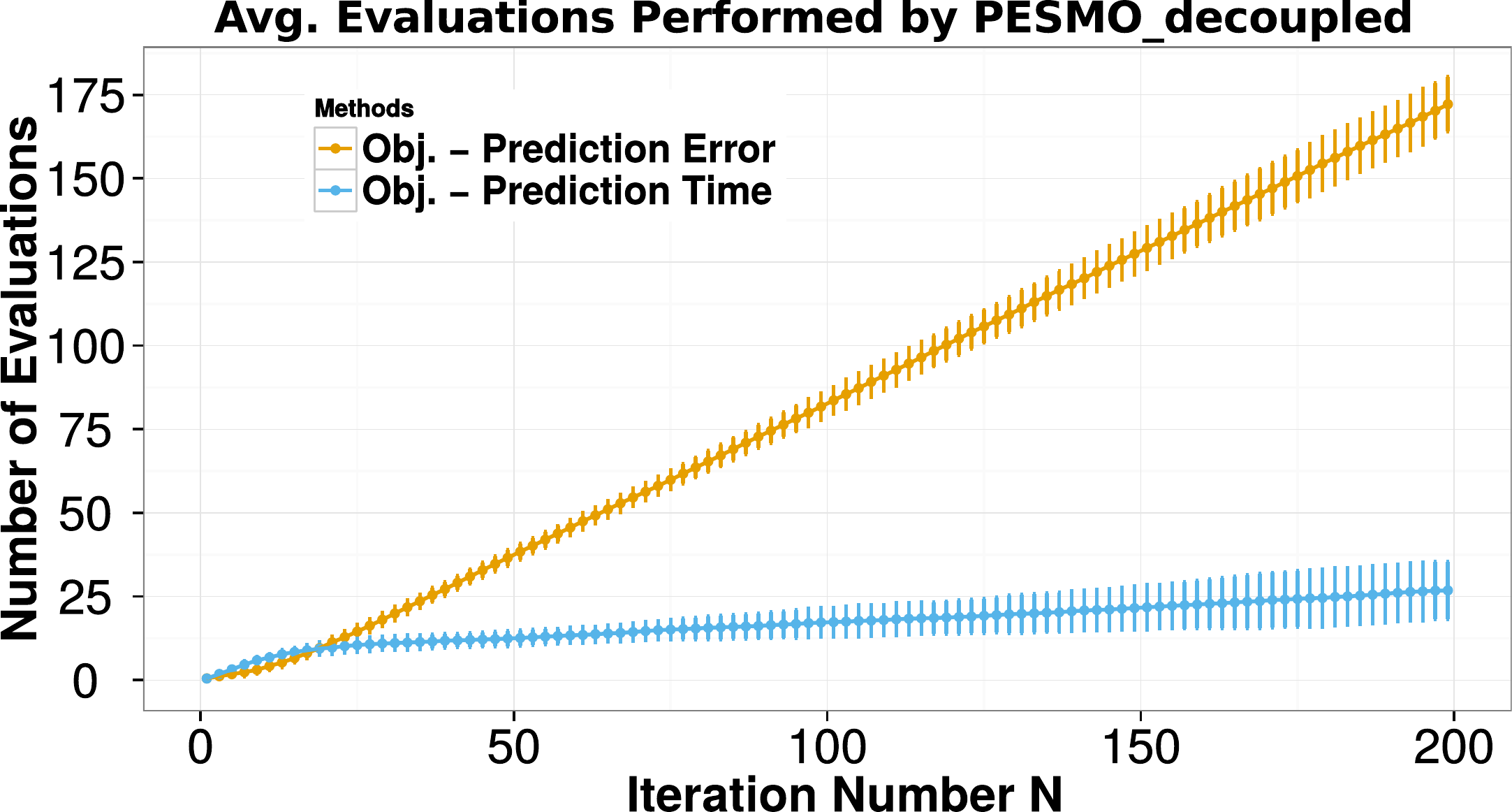}   
\end{center}
\vspace{-.35cm}
\caption{{\small Number of evaluations of each objective done by $\text{PESMO}_\text{decoupled}$,
as a function of the iteration number $N$, in the problem of finding good neural networks. Best seen in color.
}}
\label{fig:evaluations_nnet}
\vspace{-.35cm}
\end{figure}

The good results obtained by $\text{{\small PESMO}}_\text{decoupled}$ are explained by Figure  \ref{fig:evaluations_nnet}, which shows the average number of evaluations of each objective.
More precisely, the objective that measures the prediction time is evaluated just a few times.
This makes sense because it depends on only two parameters, \emph{i.e.}, the number of layers and the number of hidden units per layer. 
It is hence simpler than the prediction error. $\text{{\small PESMO}}_\text{decoupled}$ is able to detect this and focuses on the evaluation of the prediction error.  Of course, evaluating the prediction error more times is more expensive, since it involves training the neural network more times.
Nevertheless, this shows that $\text{{\small PESMO}}_\text{decoupled}$ is able to successfully discriminate between easy and difficult objective functions.

The supplementary material has extra experiments comparing each method on 
the task of finding an ensemble of decision trees of small size and good prediction 
accuracy.

\vspace{-.25cm}

\section{Conclusions}
\label{sec:conclusions}

We have described {\small PESMO}, a method for multi-objective Bayesian optimization.
At each iteration, {\small PESMO} evaluates the objective functions at the input location 
that is most expected to reduce the entropy of posterior estimate of the Pareto set.
Several synthetic experiments show that {\small PESMO} has better performance than other methods from the literature.
That is, {\small PESMO} obtains better recommendations with a smaller number of evaluations, both in  the case of noiseless and noisy observations.
Furthermore, the acquisition function of {\small PESMO} can be understood as a sum of $K$ 
individual acquisition functions, one per each of the $K$ objectives.
This allows for a \emph{decoupled} evaluation scenario, in which the most promising objective 
is identified by maximizing the individual acquisition functions.
When run in a decoupled evaluation setting, {\small PESMO} is able to identify the most difficult 
objectives and, by focusing on their evaluation, it provides better results.
This behavior of {\small PESMO} has been illustrated on a multi-objective optimization problem 
that consists of finding an accurate and fast neural network.
Finally, the computational cost of {\small PESMO} is small.
In particular, it scales linearly with the number of objectives $K$.
Other methods have an exponential cost with respect to $K$ which makes them infeasible for more than 3 objectives.

\section*{Acknowledgments}

Daniel Hern\'andez-Lobato gratefully acknowledges the use of the facilities of Centro de 
Computación Científica (CCC) at Universidad Aut\'onoma de Madrid. This author also acknowledges 
financial support from the Spanish Plan Nacional I+D+i, Grants TIN2013-42351-P and TIN2015-70308-REDT, 
and from Comunidad de Madrid, Grant S2013/ICE-2845 CASI-CAM-CM. Jos\'e Miguel Hern\'andez-Lobato acknowledges 
financial support from the Rafael del Pino Fundation. Amar Shah acknowledges support from the Qualcomm 
Innovation Fellowship program. Ryan P. Adams acknowledges support from the Alfred P. Sloan Foundation.

\bibliography{references}
\bibliographystyle{icml2016}

\appendix

\section{Detailed Description of Expectation Propagation}

In this section we describe in detail the specific steps of the EP algorithm that is required for the evaluation of the 
proposed acquisition function, PESMO. More precisely, we show how to compute the EP approximation to the conditional 
predictive distribution of each objective $f_k$.  From the main manuscript we know that that this distribution 
is obtained by multiplying the GP posteriors by the product of all the approximate factors. We also show how 
to implement the EP updates to refine each approximate factor. In our implementation we assume independence 
among the $K$ objective functions.

We assume the reader is familiar with the steps of the expectation propagation algorithm, as described in
\cite{minka2001}.

Recall from the main manuscript that all EP approximate factors $\tilde{\psi}$ are Gaussian and given by:
\begin{align}
\tilde{\psi}(\mathbf{x}_i,\mathbf{x}_j^\star) &= \prod_{k=1}^K\tilde{\phi}_k(\mathbf{x}_i,\mathbf{x}^\star_j)
\,,
\end{align}
where
\begin{align}
\tilde{\phi}_k(\mathbf{x}_i,\mathbf{x}_j^\star) &= \exp\left\{ -0.5 \left( 
	f_k(\mathbf{x}_i)^2 \tilde{v}_{i,j,k}  + 
	f_k(\mathbf{x}_j^\star)^2 \tilde{v}_{i,j,k}^\star +
	f_k(\mathbf{x}_j^\star) f_k(\mathbf{x}_i) \tilde{c}_{i,j,k} \right) \right. \nonumber \\
& \quad \left.  + 
	\tilde{m}_{i,j,k} f_k(\mathbf{x}_i) + \tilde{m}_{i,j,k}^\star f_k(\mathbf{x}_j^\star)
	\right\}
	\,,
	\label{eq:ind_factor}
\end{align}
for some input location $\mathbf{x}_i$ and for some $\mathbf{x}_j^\star$ extracted from the current sampled 
Pareto set $\mathcal{X}^\star$. Note that in (\ref{eq:ind_factor}),  $\tilde{v}_{i,k}$, $\tilde{v}_{j,k}^\star$, $\tilde{c}_{i,j,k}$,
$\tilde{m}_{i,k}$ and $\tilde{m}_{j,k}^\star$ are parameters fixed by EP.

\subsection{Reconstruction of the Conditional Predictive Distribution}
\label{sec:cond_pred_dist}

In this section we show how to obtain a conditional predictive distribution for each objective function $f_k$, 
given a sampled Pareto set $\mathcal{X}^\star=\{\mathbf{x}_1^\star,\ldots,\mathbf{x}_M^\star\}$ 
of size $M$, and a set of $N$ input locations 
$\hat{\mathcal{X}}=\{\mathbf{x}_1,\ldots,\mathbf{x}_N\}$, with corresponding 
observations of the $k$-th objective $\mathbf{y}_k$. We also assume that 
we are given the EP approximate factors $\tilde{\psi}$.

Define $\mathbf{f}_k=
(f_k(\mathbf{x}_1^\star),\ldots,f_k(\mathbf{x}_M^\star), f_k(\mathbf{x}_1),\ldots,f_k(\mathbf{x}_N))^\text{T}$. 
We are interested in computing 
\begin{align}
p(\mathbf{f}_k|\mathcal{X}^\star,\hat{\mathcal{X}},\mathbf{y}_k) & \approx
	q(\mathbf{f}_k) = 
	Z^{-1}
	p(\mathbf{f}_k|\hat{\mathcal{X}},\mathbf{y}_k) 
        \prod_{\mathbf{x} \in \mathcal{X}^\star \cup \hat{\mathcal{X}}} 
	\prod_{\substack{\mathbf{x}^\star \neq \mathbf{x} \\ \mathbf{x}^\star \in \mathcal{X}^\star}} \tilde{\phi}_k(\mathbf{x},\mathbf{x}^\star)\,,
	\label{eq:approx_cond_f_k}
\end{align}
for some normalization constant $Z$. 
In (\ref{eq:approx_cond_f_k}) we have only considered those approximate factors that depend on the current 
objective function $f_k$ and ignored the rest.
We note that $p(\mathbf{f}_k|\hat{\mathcal{X}},\mathbf{y}_k)$ is simply the posterior distribution of the Gaussian process,
which is a multi-variate Gaussian over $N+M$ variables with natural parameters $\tilde{\bm{\Sigma}}^k$ and $\tilde{\bm{\mu}}^k$. 
Furthermore, all EP approximate factors $\tilde{\psi}$ are Gaussian. Because the Gaussian distribution is closed under the
product operation, $q(\mathbf{f}_k)$ is a multi-variate Gaussian distribution over $N+M$ variables with natural parameters 
$\mathbf{S}^k$ and $\mathbf{m}^k$ obtained as:
\begin{align}
S_{i,i}^k &= \tilde{\Sigma}_{i,i}^k + \sum_{j=M+1}^{N+M} \tilde{v}_{j,i,k}^\star + 
	\sum_{j \neq i} \tilde{v}_{i,j,k} +
	\sum_{\substack{j=1 \\ j \neq i}}^M \tilde{v}_{j,i,k}^\star
\quad \text{for} \quad i=1,\ldots,M\,,
\nonumber \\
S_{i,i}^k &= \tilde{\Sigma}_{i,i}^k + \sum_{j=1}^M \tilde{v}_{i,j,k} \quad \text{for} \quad i=M+1,\ldots,N+M\,, \nonumber \\
S_{i,j}^k &= \tilde{\Sigma}_{i,j}^k + \tilde{c}_{i,j,k} \quad \text{for} \quad i=M+1,\ldots,N+M\,, \quad \text{and} \quad 
		j=1,\ldots,M\,, \nonumber \\
S_{i,j}^k &= \tilde{\Sigma}_{j,i}^k + \tilde{c}_{j,i,k}  + \tilde{c}_{i,j,k} \quad \text{for} \quad i=1,\ldots,M\,, \quad \text{and} \quad 
		j=1,\ldots,M\,, \quad \text{and} \quad i \neq j\,, \nonumber \\
S_{i,j}^k &= \tilde{\Sigma}_{i,j}^k \quad \text{for} \quad i=M+1,\ldots,N+M\,, \quad \text{and} \quad 
		j=M+1,\ldots,M+M\,, \quad \text{and} \quad i \neq j\,, \nonumber \\
S_{j,i}^k &= S_{i,j}^k \quad \text{for} \quad i\neq j\,,\nonumber\\
m_{i}^k &= \tilde{\mu}_{i}^k + \sum_{j=M+1}^{N+M} \tilde{m}_{j,i,k}^\star + 
	\sum_{j \neq i} \tilde{m}_{i,j,k} +
	\sum_{\substack{j=1 \\ j \neq i}}^M \tilde{m}_{j,i,k}^\star
\quad \text{for} \quad i=1,\ldots,M\,,
\nonumber \\
m_{i}^k &= \tilde{\mu}_{i}^k + \sum_{j=1}^M \tilde{m}_{i,j,k} \quad \text{for} \quad i=M+1,\ldots,N+M\,. 
\end{align}
From these natural parameters we can obtain, respectively, the covariance matrix $\bm{\Sigma}^k$ and the mean vector 
$\bm{\mu}^k$ by computing $(\mathbf{S}^{k})^{-1}$ and $(\mathbf{S}^{k})^{-1}\mathbf{m}^{k}$. This has a total
cost that is $\mathcal{O}((N+M)^3)$ since we have to invert a matrix of size $(N+M)\times(N+M)$.
Importantly, this operations has to be performed only once at each iteration of the optimization process, and the 
result can be reused when evaluating the acquisition at different input locations.

\subsection{The Conditional Predictive Distribution at a New Point}

Consider now the computation of the conditional distribution for $f_k$ at a new candidate location $\mathbf{x}_{N+1}$.
Assume that we have already obtained $q(\mathbf{f}_k)$ from the previous section and that we have already obtained
the parameters of the required approximate factors by using EP. We are interested in evaluating the
conditional predictive variance for $f_k(\mathbf{x}_{N+1})$. For this, we need to evaluate:
\begin{align}
p(f_k(\mathbf{x}_{N+1})|\hat{\mathcal{X}},\mathcal{X}^\star,\mathbf{x}_{N+1}) & \approx
	\int {Z}^{-1} q(\mathbf{f}_k,f_k(\mathbf{x}_{N+1})) \prod_{\mathbf{x}^\star \in \mathcal{X}^\star} 
	\tilde{\phi}_k(\mathbf{x}_{N+1},\mathbf{x}^\star)
	d \mathbf{f}_k\,,
	\label{eq:cond_for_f_k_at_x}
\end{align}
where $Z$ is simply a normalization constant and $q(\mathbf{f}_k,f_k(\mathbf{x}_{N+1}))$ is a 
multivariate Gaussian distribution which results by extending $q(\mathbf{f}_k)$ with one extra
dimension for $f_k(\mathbf{x}_{N+1})$.  Recall that $\mathbf{f}_k=
(f_k(\mathbf{x}_1^\star),\allowbreak 
\ldots,f_k(\mathbf{x}_M^\star), f_k(\mathbf{x}_1),\ldots,f_k(\mathbf{x}_N))^\text{T}$. 
Again, in (\ref{eq:cond_for_f_k_at_x}) we have only considered those
approximate factors that depend on $f_k$. The covariances between $\mathbf{f}_k$ and $f_k(\mathbf{x}_{N+1})$ 
are obtained from the GP posterior for $f_k$ given the observed data. The mean and the variance of $f_k(\mathbf{x}_{N+1})$
to be used in  $q(\mathbf{f}_k,f_k(\mathbf{x}_{N+1}))$ can also be obtained in a similar way.
Because all the factors in the r.h.s. of (\ref{eq:cond_for_f_k_at_x}) are Gaussian, the result of the integral is 
a univariate Gaussian distribution.

Define $\tilde{\mathbf{f}}_k=(f_k(\mathbf{x}_1^\star),\ldots,f_k(\mathbf{x}_M^\star), f_k(\mathbf{x}_{N+1}))^\text{T}$.
Because $\prod_{\mathbf{x}^\star \in \mathcal{X}^\star} \tilde{\phi}_k(\mathbf{x}_{N+1},\mathbf{x}^\star)$ does not 
depend on $f_k(\mathbf{x}_1),\ldots,f_k(\mathbf{x}_N)$, we can marginalize these variables 
in the r.h.s. of (\ref{eq:cond_for_f_k_at_x}) to get something proportional to:
\begin{align}
\int q(\tilde{\mathbf{f}}_k) \prod_{\mathbf{x}^\star \in \mathcal{X}^\star} \tilde{\phi}_k(\mathbf{x}_{N+1},\mathbf{x}^\star)
	\prod_{i=1}^M d f_k(\mathbf{x}_i^\star)\propto
\int \mathcal{N}(\tilde{\mathbf{f}}_k|(\mathbf{S}^x)^{-1}\mathbf{m}^x,(\mathbf{S}^x)^{-1})  \prod_{i=1}^M d f_k(\mathbf{x}_i^\star)
=
\nonumber \\
\mathcal{N}(f_k(\mathbf{x}_{N+1})|m_x, \sigma_x^2)\,,
\end{align}
where $\mathbf{m}^x$ and $\mathbf{S}^x$ are the natural parameters of the approximate conditional predictive 
distribution for $\tilde{\mathbf{f}}_k$, which is Gaussian. Similarly, $m_x$ and $\sigma_x^2$ are the mean and 
variance of the Gaussian approximation to $p(f_k(\mathbf{x}_{N+1}|\hat{\mathcal{X}},\mathcal{X}^\star,\mathbf{x}_{N+1})$.

We are interested in the evaluation of $\sigma_x^2$, which is required for entropy computation.
It is clear that $\sigma_x^2$ is given by the last diagonal entry of $(\mathbf{S}^x)^{-1}$.
In consequence, we now show how to compute $\mathbf{S}^x$ and $(\mathbf{S}^x)^{-1}_{M+1,M+1}$.
We do not give the details for computing $m_x$, because only the variance is required for the entropy computation.

Each entry in $\mathbf{S}^x$ is given by:
\begin{align}
S_{i,j}^x &= S_{i,j}^k \quad \text{for} \quad 1\leq i \leq M \quad \text{and} \quad 1\leq j \leq M\,, \quad \text{and} \quad i \neq j\,,
\nonumber \\
S_{i,j}^x &= \text{cov}(f_k(\mathbf{x}_{N+1}), f(\mathbf{x}_j^\star)) + \tilde{c}_{N+1,j,k}
	 \quad \text{for} \quad 1\leq j \leq M \quad \text{and} \quad i = M + 1\,,
	\nonumber \\
S_{j, i}^x & = S_{i, j}^x \quad \text{for} \quad j \neq i\,, \quad \text{and} \quad 1 \leq i,j \leq M\,, \nonumber \\
S_{i, i}^x & = S_{i, i}^k + \tilde{v}_{N+1,j,k}^\star\,, \text{for} \quad 1\leq i \leq M\,, \nonumber \\
S_{M+1, M+1}^x & = \text{var}(f_k(\mathbf{x}_{N+1})) + \sum_{j=1}^M \tilde{v}_{N+1,j,k}\,,
\end{align}
where $\tilde{v}_{N+1,i,k}$, $\tilde{v}_{N+1,i,k}^\star$, and 
$\tilde{c}_{N+1,i,k}$ are the parameters of each of the $M$ factors 
$\tilde{\phi}_k(\mathbf{x}_{N+1},\mathbf{x}_j^\star)$, for $j=1,\ldots,M$.
Furthermore, $\text{var}(f_k(\mathbf{x}_{N+1}))$ and $\text{cov}(f_k(\mathbf{x}_{N+1})$ 
are the posterior variance of $f_k(\mathbf{x}_{N+1})$ and the posterior covariance
between $f_k(\mathbf{x}_{N+1})$ and $f_k(\mathbf{x}_j^\star)$.

We note that $\mathbf{S}^x$ has a block structure in which only the last row and column depend on 
$\mathbf{x}_{N+1}$.  This allows to compute $\sigma_x^2=(\mathbf{S}^x)^{-1}_{M+1,M+1}$ with cost $\mathcal{O}(M^3)$
using the formulas for block matrix inversion.  All these computations are carried out using the open-BLAS library
for linear algebra operations which is particularly optimized for each processor.

Given $\sigma_x^2$ we only have to add the variance of the additive Gaussian noise $\epsilon^k_{N+1}$ to obtain the final
variance of the Gaussian approximation to the conditional predictive distribution of $y^k_{N+1}=f_k(\mathbf{x}_{N+1}) + \epsilon_{N+1}^k$.

\subsection{Update of an Approximate Factor}

EP updates until convergence each of the approximate factors $\tilde{\psi}$. 
Given an exact factor $\psi(\mathbf{x}_i,\mathbf{x}_j^\star)$, in this section we show 
how to update the corresponding EP approximate factor $\tilde{\psi}(\mathbf{x}_i,\mathbf{x}_j^\star)$. 
For this, we assume that we have already obtained the parameters $\bm{\mu}^k$ and $\bm{\Sigma}^k$ of each 
of the $K$ conditional predictive distributions, $q(\mathbf{f}_k)$, as described in Section \ref{sec:cond_pred_dist}.
The form of the exact factor is:
\begin{align}
\psi(\mathbf{x}_i,\mathbf{x}_j^\star) &= 1 - \prod_{k=1}^K\Theta \left(f_k(\mathbf{x}_j^\star) - f_k(\mathbf{x}_i)\right)\,.
\end{align}
Note that this factor only depends on $f_k(\mathbf{x}_i)$ and $f_k(\mathbf{x}_j^\star)$ for $k=1,\ldots,K$. This
means that we are only interested in the distribution of these variables under $q(\mathbf{f}_k)$, for $k=1,\ldots,K$,
and can ignore (marginalize in $q$) all other variables. Thus, in practice we will work with 
$q(f_k(\mathbf{x}_i),f_k(\mathbf{x}_j^\star))$, for $k=1,\ldots,K$. These are bi-variate Gaussian distributions. 
Let the means, variances and covariance parameters of one of these distributions 
be respectively: $m_{i,j,k}$, $m_{i,j,k}^\star$, $v_{i,j,k}$, $v_{i,j,k}^\star$ and $c_{i,j,k}$.

\subsubsection{Computation of the Cavity Distribution}

The first step of the update is to compute and old distribution $q^\text{old}$, known as the cavity distribution, which is 
obtained by removing the 
approximate factor $\tilde{\psi}(\mathbf{x}_i,\mathbf{x}_j^\star)$ from the product of the $K$ approximations 
$q(f_k(\mathbf{x}_i),f_k(\mathbf{x}_j^\star))$, for $k=1,\ldots,K$.  Recall that 
$\tilde{\psi}(\mathbf{x}_i,\mathbf{x}_j^\star) = \prod_{k=1}^K \tilde{\phi}_k(\mathbf{x}_i,\mathbf{x}_j^\star)$.
This can be done by division. Namely, 
$q^\text{old}(f_k(\mathbf{x}_i),f_k(\mathbf{x}_j^\star))\propto q(f_k(\mathbf{x}_i),f_k(\mathbf{x}_j^\star)) / 
\tilde{\phi}_k(\mathbf{x}_i,\mathbf{x}_j^\star)$, for $k=1,\ldots,K$. Because all the factors are Gaussian, the result is another
bi-variancete Gaussian distribution. Let the corresponding \emph{old} parameters be:
$m^\text{old}_{i,j,k}$, $m^{\text{old}\star}_{i,j,k}$, $v^\text{old}_{i,j,k}$, $v_{i,j,k}^{\text{old}\star}$ and 
$c^\text{old}_{i,j,k}$. These parameters are obtained by subtracting from the natural parameters of
$q(f_k(\mathbf{x}_i),f_k(\mathbf{x}_j^\star))$, the natural parameters of $\tilde{\phi}_k$. The resulting
natural parameters are then transformed into standard mean and covariance parameters to get the parameters
of $q^\text{old}(f_k(\mathbf{x}_i),f_k(\mathbf{x}_j^\star))$. This step is performed as indicated in the 
last paragraph of Section \ref{sec:cond_pred_dist}, and it involves computing the inverse 
of a $2 \times 2$ matrix, which is something very easy and inexpensive to do in practice.

\subsubsection{Computation of the Moments of the Tilted Distribution}

Given each $q^\text{old}(f_k(\mathbf{x}_i),f_k(\mathbf{x}_j^\star))$, for $k=1,\ldots,K$, 
the next step of the EP algorithm is to compute the moments of a tilted distribution defined as:
\begin{align}
\hat{p}(\{f_k(\mathbf{x}_i),f_k(\mathbf{x}_j^\star)\}_{k=1}^K) &= \hat{Z}^{-1}
	\psi(\mathbf{x}_i,\mathbf{x}_j^\star)
	\prod_{k=1}^K q^\text{old}(f_k(\mathbf{x}_i),f_k(\mathbf{x}_j^\star))\,,
\end{align}
where $\hat{Z}$ is just a normalization constant that guarantees that $\hat{p}$ integrates up to one.

Importantly, the normalization constant $\hat{Z}$  can be computed in closed form and is given by:
\begin{align}
\hat{Z} &= 1 - \prod_{k=1}^K \Phi\left(\frac{m^\text{old}_{i,j,k} - m^{\text{old}\star}_{i,j,k} }
	{\sqrt{v^\text{old}_{i,j,k} + v^{\text{old}\star}_{i,j,k} - 2 c^\text{old}_{i,j,k}}}\right)\,,
\end{align}
where $\Phi(\cdot)$ is the c.p.f. of a standard Gaussian distribution.
The moments (mean vector and covariance matrix) of $\hat{p}$ can be readily obtained 
from the derivatives of $\log \hat{Z}$ with respect to the parameters of $q^\text{old}(f_k(\mathbf{x}_i),f_k(\mathbf{x}_j^\star))$,
as indicated in the Appendix of \cite{dhernandPhd2009}.

\subsubsection{Computation of the Individual Approximate Factors}

Given the moments of the tilted distribution $\hat{p}(\{f_k(\mathbf{x}_i),f_k(\mathbf{x}_j^\star)\}_{k=1}^K)$,
it is straight-forward to obtain the parameters of the approximate factors $\tilde{\phi}_k$, for $k=1,\ldots,K$, 
whose product approximates $\psi(\mathbf{x}_i,\mathbf{x}_j^\star)$. The idea is that the product of 
$\tilde{\psi}(\mathbf{x}_i,\mathbf{x}_j^\star)=\prod_{i=1}^K\tilde{\phi}_k(\mathbf{x}_i,\mathbf{x}_j^\star)$
and $\prod_{k=1}^K q^\text{old}(f_k(\mathbf{x}_i),f_k(\mathbf{x}_j^\star))$ should lead to a Gaussian
distribution with the same moments as the tilted distribution $\hat{p}$.

The detailed steps to find each $\tilde{\phi}_k$ are: (i) Define a Gaussian distribution with the same
moments as $\hat{p}$, denoted $\prod_{k=1}^K q^\text{new}(f_k(\mathbf{x}_i),f_k(\mathbf{x}_j^\star))$.
Note that this distribution factorizes across each objective $k$. Let the parameters of this distribution
be $m^\text{new}_{i,j,k}$, $m^{\text{new}\star}_{i,j,k}$, $v^\text{new}_{i,j,k}$, $v_{i,j,k}^{\text{new}\star}$ and 
$c^\text{new}_{i,j,k}$, for $k=1,\ldots,K$. (ii) Transform these parameters to natural parameters, and subtract
to them the natural parameters of $\prod_{k=1}^K q^\text{old}(f_k(\mathbf{x}_i),f_k(\mathbf{x}_j^\star))$.
(iii) The resulting natural parameters are the natural parameters of each updated $\tilde{\phi}_k$.
Note that this operation involves going from standard parameters to natural parameters. Again, this can 
be done as indicated in the last paragraph of Section \ref{sec:cond_pred_dist}. For this, the inverse 
of the corresponding $2 \times 2$ covariance matrix of each Gaussian factor of 
$\prod_{k=1}^K q^\text{new}(f_k(\mathbf{x}_i),f_k(\mathbf{x}_j^\star))$ is required. 
Because these are $2 \times 2$ matrices, this operation is inexpensive and very fast to compute.

\subsection{Parallel EP Updates and Damping}

In our EP implementation we updated in parallel each of the approximate factors $\tilde{\psi}$,
as indicated in \cite{NIPS2009_0360}. That is, we computed the corresponding cavity distribution 
for each factor $\psi$ and updated the corresponding approximate factor $\tilde{\psi}$ afterwards. 
Next, the EP approximation was reconstructed as indicated in Section \ref{sec:cond_pred_dist}.

We also employed damped EP updates in our implementation \cite{minka2002expectation}. 
That is, the parameters of each updated factor 
are set to be a linear combination of the old parameters and the new parameters. 
The use of damped updates prevents very large changes in the parameter values. It is hence very 
useful to improve the convergence properties of the algorithm. Finally, damping does not 
change the convergence points of EP.

\section{Finding a Small and Accurate Ensemble of Decision Trees}

In this section we evaluate each of the methods from the main manuscript
in the task of finding an ensemble of decision trees of small size that has low 
prediction error. We measure the ensemble size in terms of the sum of the total number of 
nodes in each of trees of the ensemble. Note that the objectives considered are conflicting 
because it is expected that an ensemble of small size has higher prediction error than an ensemble 
of larger size. The dataset considered is the \emph{German Credit} dataset, which is extracted from 
the UCI repository \cite{Lichman:2013}. This is a binary classification dataset with 1,000 instances 
and 9 attributes.  The prediction error is measured using a 10-fold-cross validation procedure that 
is repeated 5 times to reduce the variance of the estimates.

Critically, to get ensembles of decision trees with good prediction properties one must encourage diversity 
in the ensemble \cite{dietterich2000ensemble}. In particular, if all the decision trees are equal, there 
is no gain from aggregating them in an ensemble. However, too much diversity can also lead to ensembles
of poor prediction performance. For example, if the predictions made are completely random, one cannot 
obtain improved results by aggregating the individual classifiers. In consequence, we consider here several 
mechanisms to encourage diversity in the ensemble, and let the amount of diversity be specified in terms 
of adjustable parameters.

To build the ensemble we employed decision trees in which the data is split at each 
node, and the best split is chosen by considering each time a random set of attributes ---we use the 
\emph{Decision-Tree} implementation provided in the python package \emph{scikit-learn} for this,
and the number of random attributes is an adjustable parameter. This is
the approach followed in Random Forest \cite{breiman01random} to generate the ensemble classifiers. 
Each tree is trained on a random subset of the training data of a particular size, which is another 
adjustable parameter. This approach is known in the literature as subbagging \cite{bulman2001}, 
and has been shown to lead to classification ensembles with 
good prediction properties. We consider also an extra method to introduce diversity known as class-switching 
\cite{Martinez-Munoz2005}. In class-switching, the labels of a random fraction of the training data are changed to 
a different class.  The final ensemble prediction is computed by majority voting.

In summary, the adjustable parameters are: the number of decision trees built (between $1$ and $1,000$), the number 
of random features considered at each split in the building process of each tree (between $1$ and $9$), the minimum 
number of samples required to split a node (between $2$ and $200$), the fraction of randomly selected training data 
used to build each tree, and the fraction of training instances whose labels are changed (after doing the sub-sampling process).

Finally, we note that this setting is suited to the decoupled version of PESMO since
both objectives can be evaluated separately. In particular, the total number of nodes is estimated 
by building only once the ensemble without leaving any data aside for validation, as opposed to the
cross-validation approach used to estimate the ensemble error, which requires to build several ensembles 
on subsets of the data, to then estimate the prediction error on the data left out for validation.

We run each method for $200$ evaluations of the objectives and report results after $100$ and $200$ 
evaluations. That is, after $100$ and $200$ evaluations, we optimize the posterior means of the GPs and provide
a recommendation in the form of a Pareto set. As in the experiments reported in the main manuscript with neural 
networks, we re-estimate three times the objectives associated to each Pareto point from the recommendation made 
by each method, and average results. The goal of this averaging process is to reduce the noise in the final evaluation 
of the objectives. These final evaluations are used to estimate the performance of each method using the 
hyper-volume. We repeat these experiments $50$ times and report the average results across repetitions.

\begin{table}[htb]
\caption{{\small Avg. hyper-volume after $100$ and $200$ evaluations of the objectives.}}
\label{tab:ensemble}
\begin{center}
\begin{tabular}{@{\hspace{0mm}}c@{\hspace{2.5mm}}r@{$\pm$}l@{\hspace{2.5mm}}r@{$\pm$}l@{\hspace{2.5mm}}
	r@{$\pm$}l@{\hspace{2.5mm}}r@{$\pm$}l@{\hspace{2.5mm}}r@{$\pm$}l@{\hspace{2.5mm}}r@{$\pm$}l@{\hspace{0mm}}}
\hline
{\bf \# Eval.} & 
\multicolumn{2}{@{\hspace{2.5mm}}c@{\hspace{2.5mm}}}{\bf PESMO} &
\multicolumn{2}{@{\hspace{2.5mm}}c@{\hspace{2.5mm}}}{\bf $\text{PESMO}_\text{dec}$} &
\multicolumn{2}{@{\hspace{0mm}}c@{\hspace{0mm}}}{\bf ParEGO} &
\multicolumn{2}{@{\hspace{2.5mm}}c@{\hspace{2.5mm}}}{\bf SMSego} &
\multicolumn{2}{@{\hspace{2.5mm}}c@{\hspace{2.5mm}}}{\bf EHI} &
\multicolumn{2}{@{\hspace{2.5mm}}c@{\hspace{2.5mm}}}{\bf SUR} \\
\hline
100 &\hspace{2mm} 8.742 & .006 & \hspace{1mm}{\bf 8.755} & {\bf .009} & 8.662 & .019 & \hspace{2mm} 8.719 & .012 & 8.731 & .009 & 8.739 & .007 \\
\hline
200 & \hspace{2mm}{\bf 8.764} & {\bf .007} &\hspace{1mm} 8.758 & .007 & 8.705 & .008 &\hspace{2mm}  8.742 & .006 & 
	8.727 & .008 & 8.756 & .006 \\
\hline
\end{tabular}
\end{center}
\end{table}

Table \ref{tab:ensemble} shows the average hyper-volume of the recommendations made by each method,
after 100 and 200 evaluations of the objective functions. The table also shows the corresponding error bars.
In this case the observed differences among the different methods are smaller than in the experiments with 
neural networks. Nevertheless, we observe that the decoupled version of PESMO obtains the best results after 100 evaluations. After this, 
PESMO in the coupled setting performs best, closely followed by SUR. After 200 evaluations, the best 
method is the coupled version of PESMO, closely followed by its decoupled version and by SUR. SMSego and 
EHI give worse results than these methods, in general. Finally, as in the experiments with neural networks 
reported in the main manuscript, ParEGO is the worst performing method. In summary, the best methods 
are PESMO in either setting (coupled or decoupled) and SUR. All other methods perform worse.
Furthermore, the decoupled version of PESMO gives slightly better results at the beginning, \emph{i.e.}, after 
100 evaluations.

\begin{figure}[hbt]
\begin{center}
\begin{tabular}{cc}
\includegraphics[width = 0.77 \textwidth]{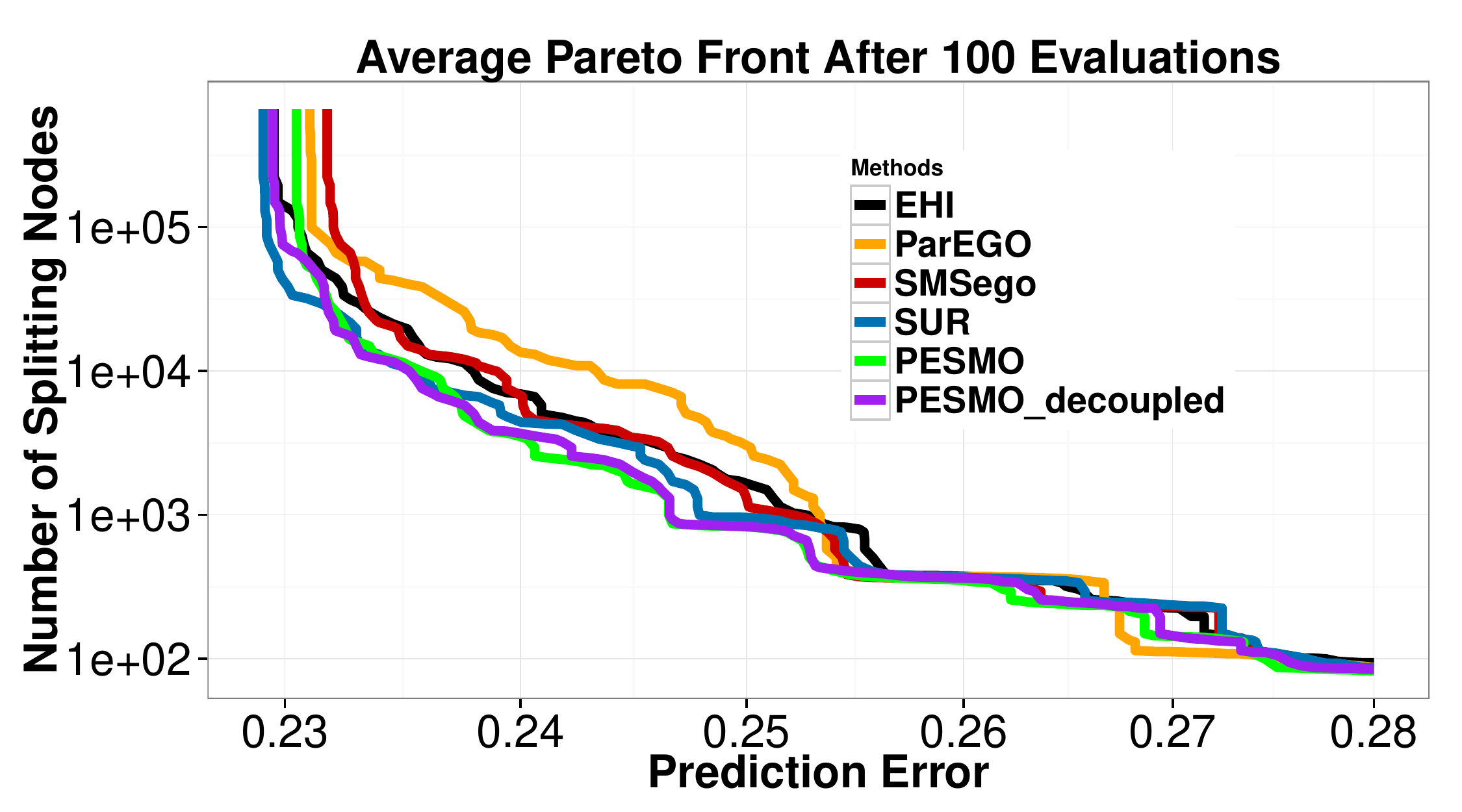}  \\
\includegraphics[width = 0.77 \textwidth]{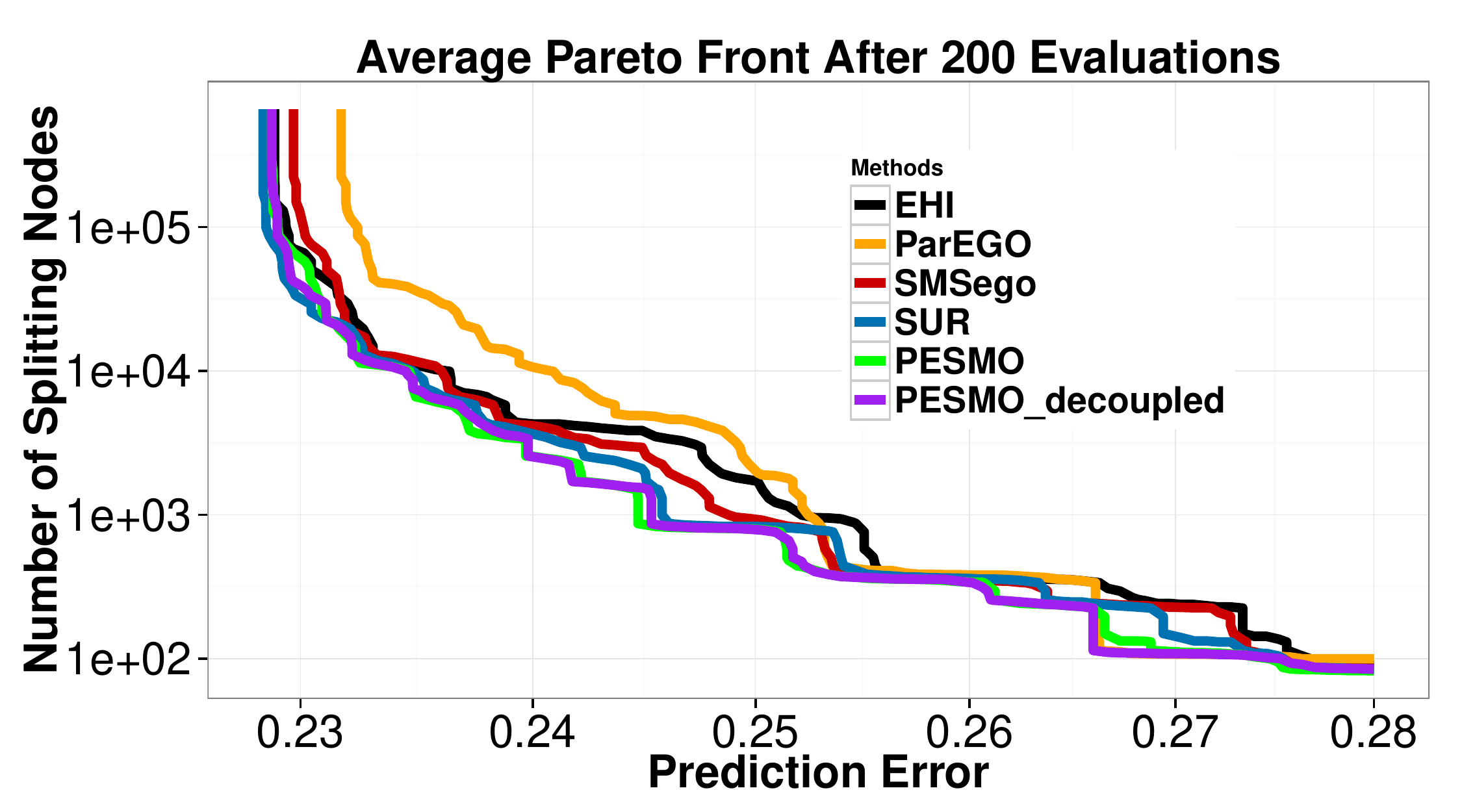}  
\end{tabular}
\end{center}
\caption{{\small Avg. Pareto fronts obtained by each method after 
100 (top) and 200 (bottom) evaluations of the objectives. Best seen in color. 
}}
\label{fig:frontiers_ensemble}
\end{figure}

Figure \ref{fig:frontiers_ensemble} shows the average Pareto front (this is simply the values in functional 
space associated to the Pareto set) corresponding to the recommendations made
by each method after 100 (top) and 200 evaluations of the objectives (bottom). We observe that 
PESMO finds ensembles with better properties than the ones found by EHI, SMSego and ParEGO.
Namely, ensembles of smaller size for a similar or even better prediction error.
The most accurate ensembles are found by SUR. Nevertheless, they have a very similar error to the 
one of the most accurate ensembles found by PESMO. Finally, we note that in some cases, PESMO is able 
to find ensembles of intermediate size with better prediction error than the ones found by SUR.

\begin{figure}[htb]
\begin{center}
\includegraphics[width = 0.77 \textwidth]{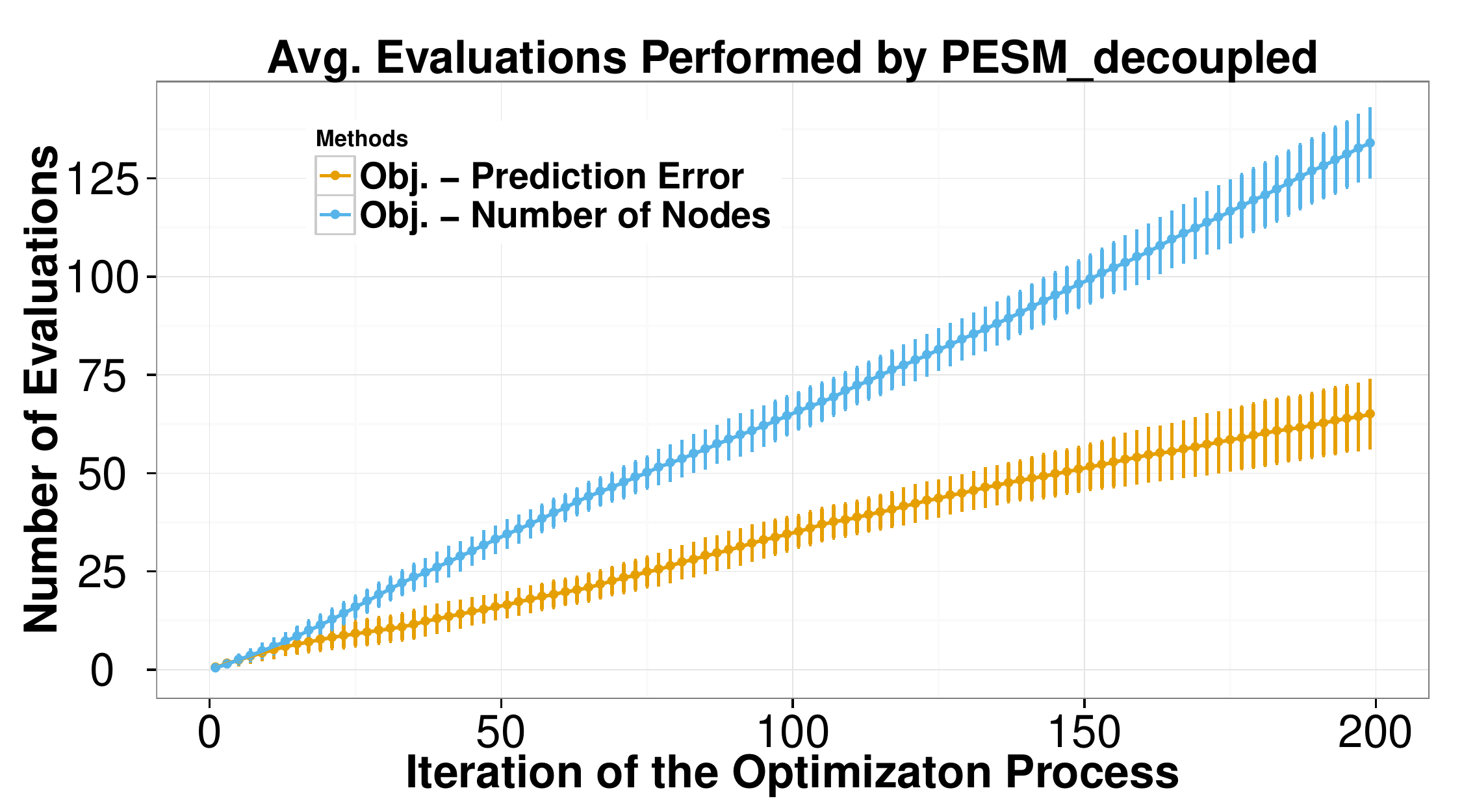}   
\end{center}
\caption{{\small Number of evaluations of each objective done by $\text{PESMO}_\text{decoupled}$,
as a function of the iteration number $N$, in the problem of finding a good ensemble of decision trees. Best seen in color.
}}
\label{fig:evaluations_ensemble}
\end{figure}

Figure \ref{fig:evaluations_ensemble} shows the average number of times that the decoupled
version of PESMO evaluates each objective. We observe that in this case the objective that
measures the number of nodes in the ensemble is evaluated more times. However, the difference
between the number of evaluations of each objective  is smaller than the difference observed 
in the case of the experiments with neural networks. Namely, 135 evaluations of one objective 
versus 65 evaluations of the other, in this case, compared to 175 evaluations versus 25 evaluations, 
in the case of the experiments with neural networks.  This may explain why in this case the differences 
between the coupled and the decoupled version of PESMO are not as big as in the experiments reported 
in the main manuscript.

\section{Accuracy of the Acquisition in the Decoupled Setting}

One question to be experimentally addressed is whether the proposed approximations for the individual acquisition 
functions $\alpha_k(\cdot)$, for $k=1,\ldots,K$, with $K$ the total number of objectives are sufficiently 
accurate in the decoupled case of PESMO. For this, we extend the experiment carried out in the main manuscript, and 
compare in a one-dimensional problem with two objectives the acquisition functions $\alpha_1(\cdot)$ and 
$\alpha_2(\cdot)$ computed by {\small PESMO}, with a more accurate estimate obtained 
via expensive Monte Carlo sampling and a non-parametric estimator of the entropy \cite{singh2003nearest}.
This estimate measures the expected decrease in the entropy of the predictive distribution of one of the objectives,
at a given location of the input space, after conditioning to the Pareto set.
Importantly, in the decoupled case, the observations corresponding to each objective need not be located at 
the same input locations.

\begin{figure}[h!]
\begin{center}
\begin{tabular}{cc}
\includegraphics[width = 0.75 \textwidth]{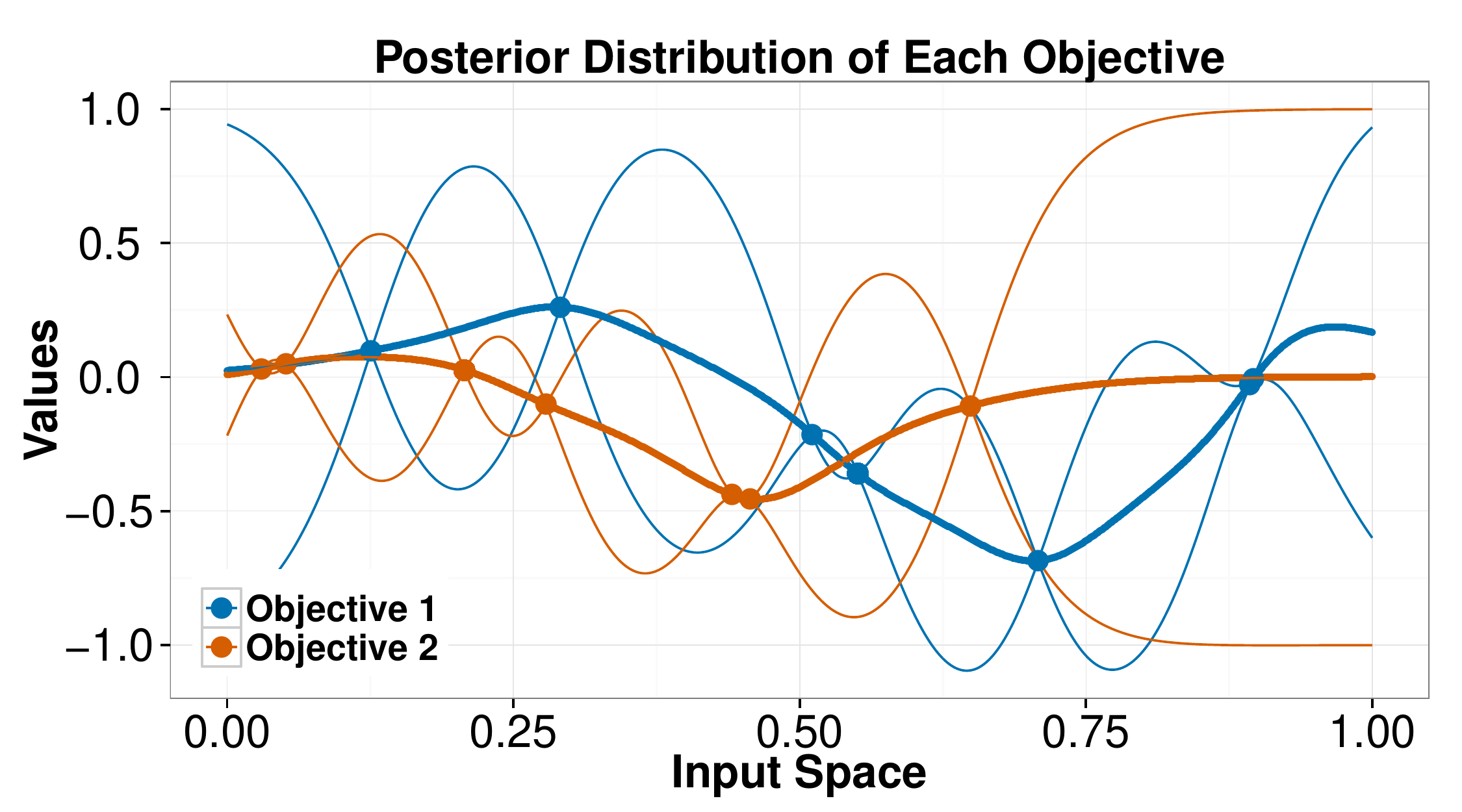}  \\
\includegraphics[width = 0.75 \textwidth]{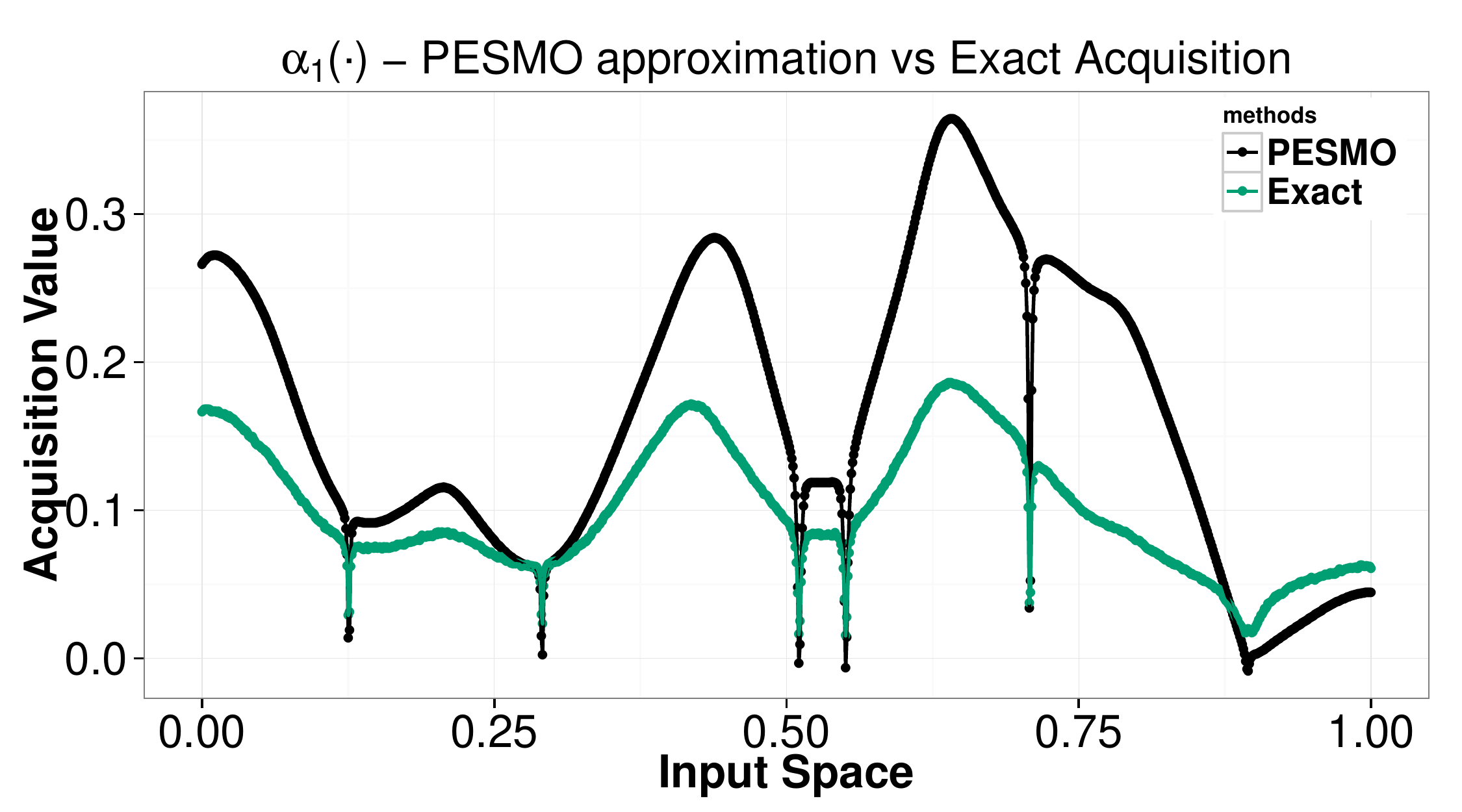}  \\
\includegraphics[width = 0.75 \textwidth]{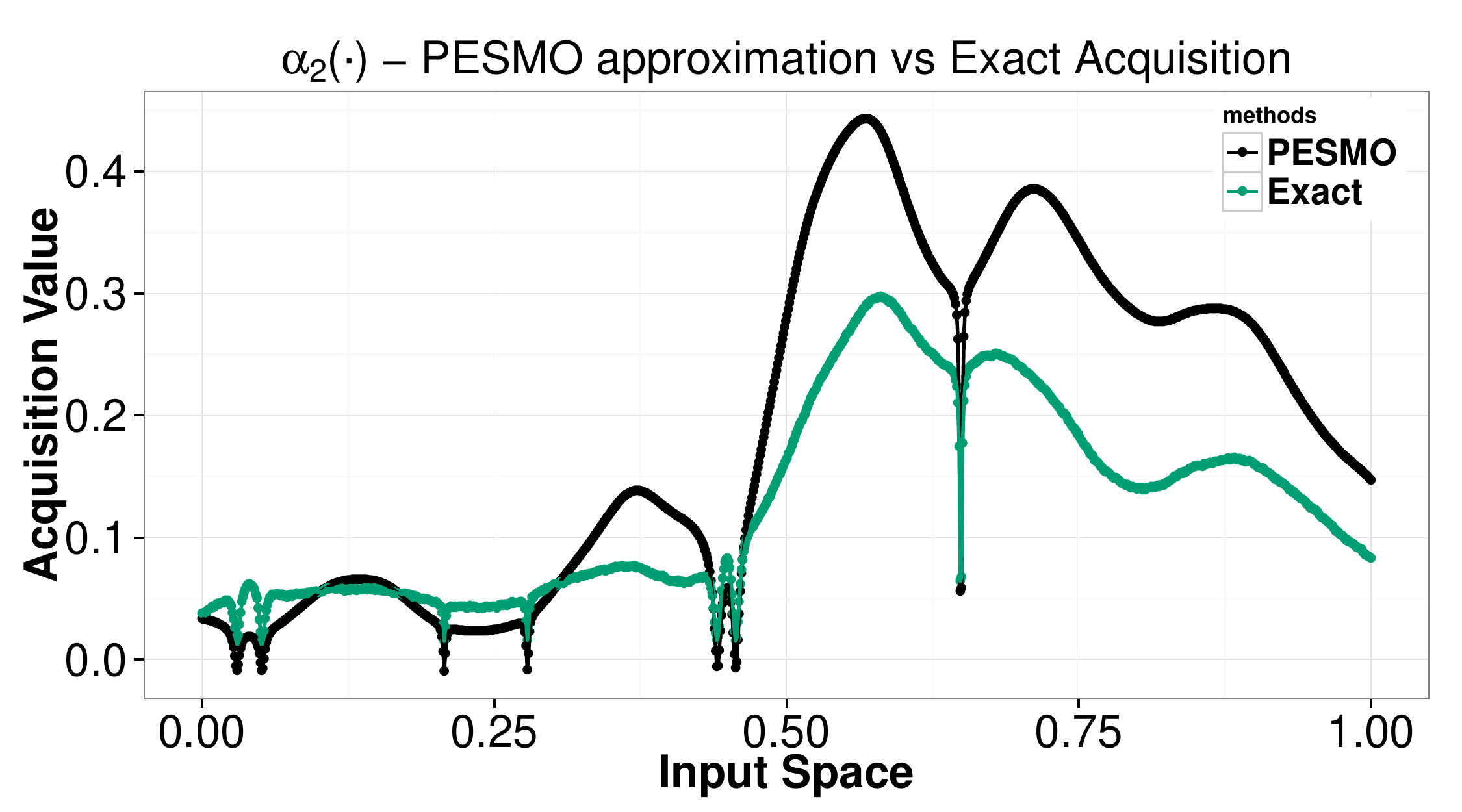}  
\end{tabular}
\end{center}
\caption{{\small (top) Observations of each objective and posterior mean and standard deviations of each 
GP model. (middle) Estimates of the acquisition function corresponding to the first objective, $\alpha_1(\dot)$, by {\small PESMO}, and by
a Monte Carlo method combined with a non-parametric estimator of the entropy. (bottom) Same results for the acquisition
function corresponding to the second objective $\alpha_2(\cdot)$. 
Best seen in color.}}
\label{fig:acq_dec}
\end{figure}

Figure \ref{fig:acq_dec} (top) shows at a given step of the optimization process, the observed data and the posterior mean 
and the standard deviation of each of the two objectives. The figure on the middle shows the corresponding acquisition function 
corresponding to the first objective, $\alpha_1(\cdot)$, computed by {\small PESMO} and by the Monte Carlo method (Exact). 
The figure on the bottom shows the same results for the acquisition function corresponding to the second objective, $\alpha_2(\cdot)$.
We observe that both functions look very similar, including the location of the global maximizer. This indicates that the 
approximation obtained by expectation propagation is potentially good also in the decoupled setting.

\end{document}